\documentclass[11pt,a4paper]{article}
\usepackage[hyperref]{acl2021}
\usepackage{times}
\usepackage{graphicx}
\usepackage{subcaption}
\usepackage{latexsym}
\usepackage{booktabs}
\usepackage{enumitem}

\pdfoutput=1

\usepackage{microtype}

\title{On Positivity Bias in Negative Reviews}

\author{Madhusudhan Aithal \\
  University of Colorado Boulder \\
  \texttt{madhuaithal@colorado.edu} \\\And
  Chenhao Tan \\
  University of Chicago \\
  \texttt{chenhao@uchicago.edu} \\}

\date{}

\aclfinalcopy

\newif\ifhidecomments
\hidecommentsfalse

\ifhidecomments
    \newcommand{\madhu}[1]{}
    \newcommand{\chenhao}[1]{}
\else
    \newcommand{\chenhao}[1]{\textcolor{red}{\textsc{\textbf{[#1 --ct]}}}}
    \newcommand{\madhu}[1]{\textcolor{blue}{\textsc{\textbf{[#1 --ma]}}}}
\fi

\newcommand{\para}[1]{\smallskip\noindent{\bf #1}}
\newcommand{\figref}[1]{Figure~\ref{#1}}

\newcommand{\tableref}[1]{Table~\ref{#1}}

\begin{document}
\maketitle
\begin{abstract}
Prior work has revealed that positive words occur more frequently than negative words in human expressions, which is typically attributed to positivity bias, a tendency for people to report positive views of reality.
But what about the language used in negative reviews?
Consistent with prior work, we show that 
English negative reviews tend to contain more positive words than negative words,
using a %
variety of datasets.
We reconcile this observation with previous findings on the pragmatics of negation, and
show that negations are commonly associated with positive words in negative reviews.
Furthermore, in negative reviews, the majority of sentences with positive words express negative opinions based on sentiment classifiers, indicating some form of negation.
\end{abstract}

\section{Introduction}

A battery of studies have validated the Pollyanna hypothesis that positive words occur more frequently than negative words in human expressions, using corpora ranging from Google Books to Twitter \citep{dodds2015human,garcia2012positive,boucher1969pollyanna,kloumann2012positivity}.
The typical interpretation is connected with the positivity bias, which broadly denotes 1) a tendency for people to report positive views of reality, 2) a tendency to hold positive expectations, views, and memories, and 3) a tendency to favor positive information in reasoning \citep{carr2011positive,augustine_positivity_2011,hoorens_positivity_2014}.
However, it remains an open question whether the Pollyanna hypothesis holds in negative reviews, where the communicative goal is to express negative opinions.

In this work, we use a wide variety of review datasets to examine the use of positive and negative words in negative reviews.
\tableref{tab:example} shows a negative review from Yelp.
Although the overall opinion is clearly negative, the author expressed the excitement to try the place and deemed the food OK.
Zooming into individual words, they used the same number of positive and negative words in this negative review.
Interestingly, this short review has as many as three negations, one directly applied to ``great'' (hence ``not great'').
More generally, we find that 
negative reviews contain {\em more} positive words than negative words,
which is consistent with the Pollyanna hypothesis.
Two possible reasons may explain this observation:
1) negative reviews tend to still include positive opinions 
due to a na\"ive interpretation of the positivity bias, where positive words express positive sentiments without accounting for negation or other contextual meaning of these words;
2) negative reviews tend to use {\em indirect expressions} (i.e., applying negations to positive words) to indicate negative opinions (e.g., ``not clean'').
Note that a broad interpretation of positivity bias may encompass the second reason,\footnote{\citet{boucher1969pollyanna} used a morphological analysis to show negative affixes are more commonly applied to positive words than negative words (unhappy vs. non-violent).} but indirect expressions could also be related to other factors, e.g., verbal politeness \citep{brown1987politeness}).

\begin{table}[t]
\centering
\begin{tabular}{ p{0.45\textwidth} } 
\toprule
Food was \textcolor{blue}{ok}...{\em not} the money they charge. I was \textcolor{red}{unimpressed} and will {\em not} return.   I was \textcolor{blue}{excited} to try this place and was so \textcolor{red}{disappointed} as my expectations were high. Service {\em not} \textcolor{blue}{great} and The parking is \textcolor{red}{awful}.\\
\bottomrule
\end{tabular}
\caption{Example negative review on Yelp. 
Positive words are in blue and negative words are in red, based on  Vader \citep{hutto2014vader}.
Negations are in italics.
This short review contains three negations.
}
\label{tab:example}
\end{table}

We aim to delineate 
these two reasons by examining the use of negations.
Our results provide support for the latter reason: negative reviews tend to use more negations than positive reviews.
The differences become even more salient when we compare negations applied to positive words vs. negative words.
Finally, among sentences with positive words in negative reviews, the majority are classified as negative than as positive by sentiment classifiers, indicating some form of negation.

\section{Related Work}

In addition to positivity bias, 
our work is closely related to experimental studies on understanding the effect of direct (e.g., ``bad'') and indirect (e.g., ``not good'') wordings.
\citet{colston1999not} and \citet{kamoen2015hotel} 
observe no difference in people's interpretation of direct 
and indirect 
wordings in negative opinions; but direct wordings receive higher evaluations than indirect ones in positive opinions.
In this work, we examine whether and how individuals use indirect wordings {\em in practice} (in negative reviews).

Our work is also related to \citet{potts2010negativity},
which finds that negation is used more frequently in negative reviews and is thus pragmatically negative. 
We extend \citet{potts2010negativity} in two ways: 
1) we demonstrate a high frequency of negation followed by positive words in negative reviews compared to other combinations, a new observation motivated 
through the lens of positivity bias;
2) we conduct a systematic study 
using a wide variety of datasets with multiple dictionaries.

Finally, our work builds on sentiment classification \citep{pang2002thumbs,pang2008opinion,liu2012sentiment}.
The NLP community has made significant progress in recognizing the sentiment in texts of various languages, obtaining accuracies of over 95\% (English) in binary classification \citep{devlin2018bert,liu2019roberta}.
Researchers have also developed novel approaches to identify fine-grained sentiments (e.g., aspect-level sentiment analysis \citep{schouten2015survey,wang2016attention,yang2013joint}) as well as semi-supervised and unsupervised approaches \citep{hu2013unsupervised,zhou2010active,tan2011user}.

\section{Datasets}

We use a wide range of English review datasets to ensure that our results are robust 
across domains.

\begin{itemize}[itemsep=-4pt,leftmargin=*,topsep=0pt]
    \item Yelp.\footnote{\url{https://www.yelp.com/dataset}.}
    We only consider restaurant reviews.
    \item IMDB movie reviews \cite{maas2011learning}.
    This dataset provides train and test splits, so we follow their split when appropriate.
    \item Stanford sentiment treebank (SST) \cite{socher2013recursive}. SST contains processed snippets of reviews from the Rotten Tomatoes website (movie reviews). It has ground truth sentiment scores of reviews at the sentence level and the word level. 
    \item Tripadvisor \cite{wang2010latent}.
    This dataset consists of hotel reviews.
    \item PeerRead \cite{kang2018dataset}. 
    We use reviews for papers in ACL, CoNLL, and ICLR.

    \item Amazon \cite{ni2019justifying}. This dataset contains Amazon reviews %
    grouped by categories. 
    We choose five categories that are substantially different from movies, hotels, and restaurants to ensure that our results are robust,
namely, ``Automotive'',
    ``Cellphones and accessories'',  ``Luxury beauty'', ``Pet supplies'', and ``Sports and outdoors''. 
\end{itemize}

\begin{figure}[t]
\centering
\includegraphics[width=0.45\textwidth]{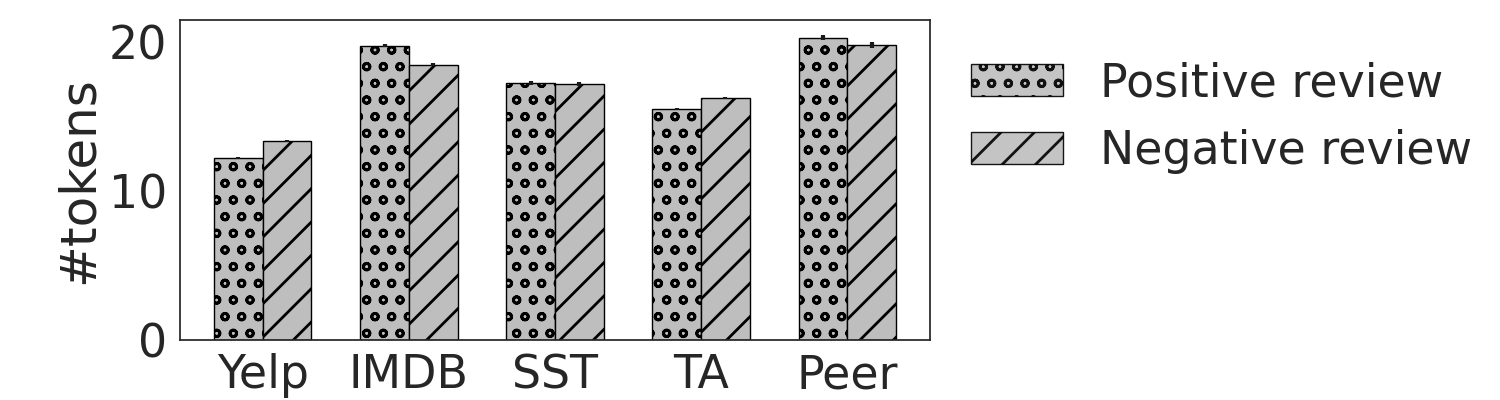}
\caption{Sentence-length comparison. 
Although negative reviews can be 
much 
longer than positive reviews, sentences in positive reviews and negative reviews have similar lengths. Results on Amazon reviews are shown in the appendix. 
Tiny error bars show standard errors.
}
\label{fig:sentence_length_cmp_non_amz}
\end{figure}
For datasets with ratings in 1-5 scale, we label reviews with ratings greater than 3 as positive and reviews with ratings less than 3 as negative following prior work \citep{pang2002thumbs}, and ignore reviews with rating 3. Similarly, for datasets with ratings scale of 1-10 (IMDB, ICLR reviews in PeerRead 
), we label reviews with ratings greater than 6 as positive and review with ratings less than 5 as negative, and ignore reviews with ratings 5 and 6.

We use spaCy
to tokenize the reviews in all datasets \citep{spacy2}, except that 
Stanford Core NLP is used to tokenize SST reviews \citep{manning2014stanford}.
We present results for Amazon reviews in the appendix,
and our main results are robust on Amazon reviews.
Our code is available at \href{https://github.com/madhu-aithal/Positivity-Bias-in-Negative-Reviews}{https://github.com/madhu-aithal/Positivity-Bias-in-Negative-Reviews}. 

\para{Length of positive vs. negative opinions.}
In general, negative opinions tend to be longer than positive opinions ($p < 0.05$ after Bonferroni correction in 6 out 10 datasets; see the appendix for details).
In comparison, the difference in length is smaller at the sentence level (\figref{fig:sentence_length_cmp_non_amz}).
Therefore, we use sentences as the basic unit in this work.
To further rule out sentence length as a 
confounding factor, we 
also 
present word-level results in the appendix. 

\section{Results}

We first 
investigate the occurrences of positive words, negative words, and negations in reviews.
We find that negative reviews contain more positive words than negative words 
in all datasets.
We show that this observation relates to the 
prevalence of negation in negative reviews compared to positive reviews in all datasets.
Furthermore, these negations are commonly associated with positive words in all datasets,
and sentences with positive words tend to be negative based on sentence-level prediction, supporting the prevalence of indirect wordings in negative reviews.

\subsection{Negative Reviews Have More Positive Words than Negative Words}
We use lexicon-based methods to examine the frequency of positive and negative words in reviews. 
For most of the datasets, 
we randomly sample 5,000 positive reviews and 5,000 negative reviews 
to compute the lexicon distribution using LIWC \citep{pennebaker2007linguistic} and Vader \citep{hutto2014vader}. In the case of SST, PeerRead, and negative reviews of Amazon Luxury Beauty, we use the entire dataset for our analysis as it has a relatively small number of reviews.

\figref{fig:vader_pos_neg_non_amz} shows that as expected, negative reviews have more negative words and fewer positive words than positive reviews, based on Vader.
Intriguingly, despite the negative nature of  negative reviews, they tend to have more {\em positive} words than {\em negative} words ($p < 0.001$ on all datasets except SST after Bonferroni correction).
Our results are robust at the word level and also hold based on LIWC and validate the Pollyanna hypothesis even in negative reviews.

\subsection{Negative Reviews Have More Negations and Indirect Expressions}

We hypothesize that in addition to the tendency to report positive views of reality, an important factor that can explain this observation in negative reviews is the use of indirect expressions (i.e., negation of positive words).
To measure the amount of negation, we use two 
approaches:
1) a lexicon-driven approach based on Vader including {\em aint}, {\em cannot}, {\em not}, and {\em never} \citep{hutto2014vader}\footnote{See the appendix for the full list of negation lexicons.};
2) the negation relation in dependency parsing.\footnote{We used spaCy for dependency parsing \citep{spacy2}.}
We present the results based on Vader negation in the main paper as it may have higher precision, and all results hold using dependency parsing.

\begin{figure}
\centering
  \includegraphics[width=0.45\textwidth]{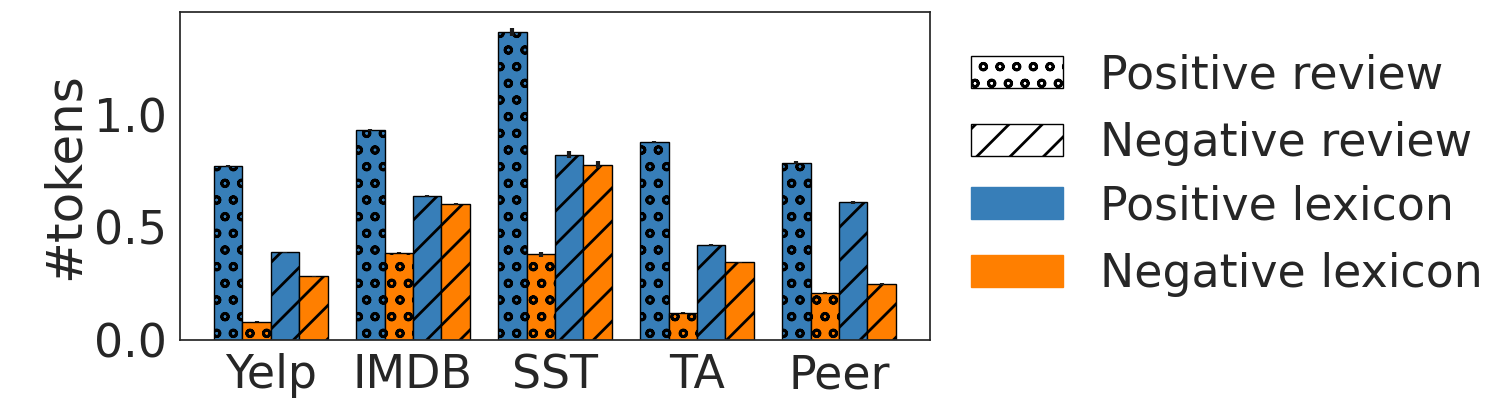}
  \caption{Number of positive and negative words based on Vader. Negative reviews have more positive words than negative words.
  }
  \label{fig:vader_pos_neg_non_amz}
\end{figure}

\begin{figure}
\centering
\begin{subfigure}[t]{0.45\textwidth}
  \centering
  \includegraphics[width=\textwidth]{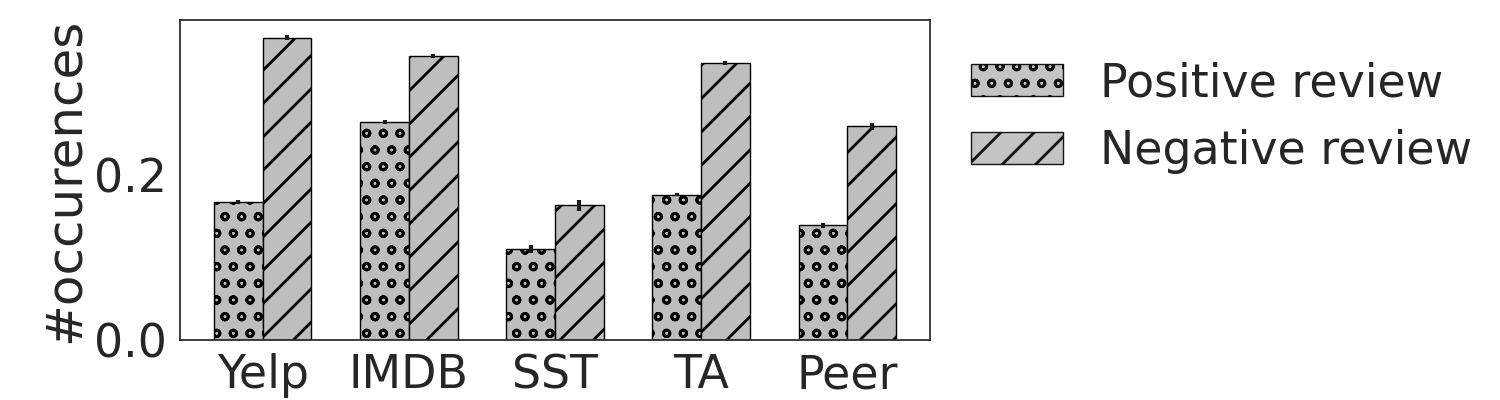}
  \caption{Overall negation.
  }
  \label{fig:overall_negation_non_amz}
\end{subfigure}
\begin{subfigure}[t]{.45\textwidth}
  \centering
  \includegraphics[width=\textwidth]{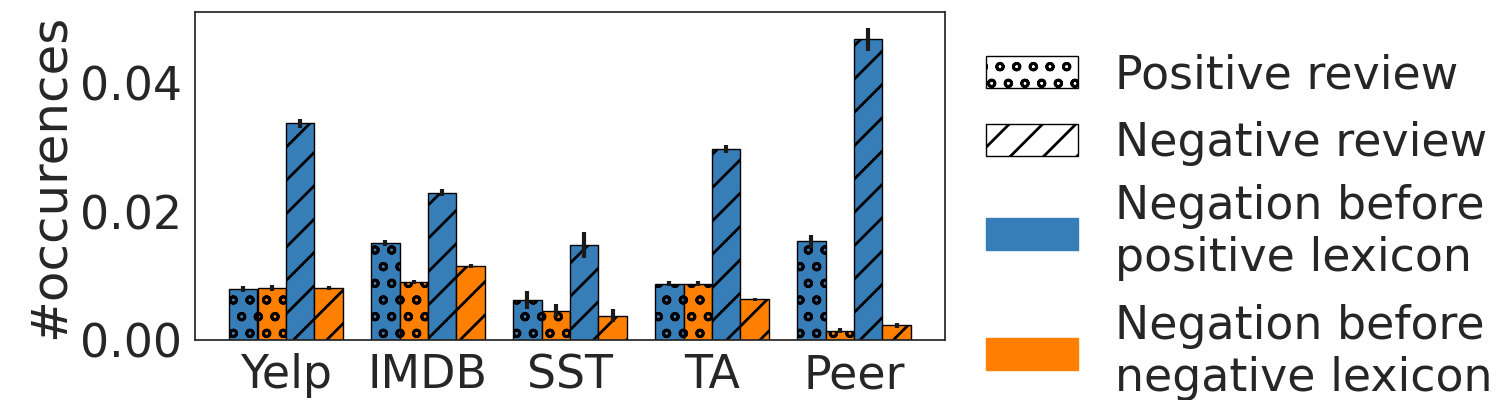}
  \caption{Negation before positive and negative lexicons.
  }
  \label{fig:sentiment_negation_non_amz}
\end{subfigure}
\caption{Negative reviews generally have more negations at the sentence level (\figref{fig:overall_negation_non_amz}).
Among those negations, \figref{fig:sentiment_negation_non_amz} shows  that there are substantially more negations before positive lexicons in negative reviews than any other combinations.
}
\label{fig:negation}
\end{figure}

\para{Negative reviews have more negations than positive reviews in all datasets.}
\figref{fig:overall_negation_non_amz} presents the number of negations at the sentence level.
In all datasets, negative reviews have more negations than positive reviews ($p < 0.001$ in all datasets). In fact, the number of negations in negative reviews almost doubles that in positive reviews in Yelp, TripAdvisor, and PeerRead (see samples in the appendix).
This observation is robust at the word level, which accounts for the fact that negative reviews tend to be longer.

\para{Negations are commonly associated with positive words in negative reviews.}
To further examine the relation between negations and sentimental lexicons, we investigate the occurrences of negations immediately followed by positive words and negative words. 
\figref{fig:sentiment_negation_non_amz} shows that there are more negations before positive words in negative reviews than any other combination ($p < 0.001$ in all datasets).
The difference is especially salient in Yelp, TripAdvisor, and PeerRead.
In particular, in negative reviews in PeerRead, negation before positive lexicon are approximately 20
times as frequent as negation before negative lexicon.
These results demonstrate the prevalence of indirect wordings when people express negative opinions.
Moreover, using indirect expressions to express negative opinions (negation before positive words) is also common in positive reviews for IMDB and PeerRead. 

\tableref{tab:common_negation_pos_words} shows the 20 most common words %
that immediately follow negations in Yelp, IMDB, and PeerRead, highlighting the prevalence of ``not clear'', ``not convincing'', and ``not surprising'' in negative reviews of NLP/ML submissions.

\begin{table}[t]
  \small
  \centering
  \begin{tabular}{l@{\hspace{4pt}}p{0.37\textwidth}}
    \toprule
    Dataset & Positive words associated with negations \\
    \midrule
    Yelp & recommend, sure, like, good, care, great, special, impressed, fresh, help, ready, enjoy, friendly, honor, helpful, clean, happy, accept, greeted, amazing \\
    \midrule
    IMDB & like, care, funny, help, sure, recommend, good, save, fit, great, special, interesting, enjoy, well, play, better, giving, original, convincing, true \\
    \midrule
    PeerRead & clear, sure, convincing, convinced, ready, well, true, clearly, surprising, novel, convincingly, recommend, guarantee, improve, interesting, support, satisfactory, help, acceptable, convince \\
    \bottomrule
  \end{tabular}
  \caption{Most frequent positive words that immediately follow negations in negative reviews, based on Vader.}
  \label{tab:common_negation_pos_words}
\end{table}

A natural question is how much of the 
usage of positive words in \figref{fig:vader_pos_neg_non_amz} can be explained by negations before positive words. We find that it is sufficient to explain 11.3\% on average. 
For instance, negative reviews in Yelp have 0.389 positive words per sentence, out of which 0.033 words follow a negation. This accounts for 
8.7\% of the usage of positive words.
This suggests that negations before positive words only account for a small fraction of positive words, despite that they dominate other combinations of negations and sentiment lexicon.
We hypothesize for positive words in negative reviews, they may be negated in ways beyond immediate preceding negations (e.g., ``nor is the food great'' and ``fail to support'').

Similarly, the number of negations followed by positive/negative words is a 
fraction of all the negations (14.2\% in negative reviews 
and 9.7\% in positive reviews).
For example, ``I will not return'' counts as negation but there is no sentimental lexicon.
We hypothesize that these negations also tend to express negative sentiments.

\subsection{Sentence-level Sentiment Classification}

To capture the sentiment of sentences with 
positive words or negations beyond negations immediately followed by positive words,
we rely on sentiment classifiers.
Specifically, we use sentence-level classification to quantify the extent of negative sentences in those contexts compared to the overall average in negative reviews.

We fine-tune BERT \citep{devlin2018bert} to perform review-level classification  for each dataset except SST and PeerRead. This is because all reviews in SST are very short and sentences in negative reviews are mostly negative whether negation occurs or not. In the case of PeerRead, the number of samples is too small to fine-tune the BERT model. 
For all other datasets except IMDB and Amazon Luxury Beauty, we randomly sample 25K positive reviews and 25K negative reviews as the training set, and 5K positive reviews and 5K negative reviews as the test set. For IMDB, we use 12.5K positive and 12.5K negative training samples provided for fine-tuning, and %
for Luxury Beauty, we use a balanced set of 2.3K positive and 2.3K negative samples for fine-tuning.
We use 90\% of the training samples 
to fine-tune the BERT model and 10\% as the development set to select hyperparameters. 
We achieved accuracies varying from 94\% to 98\% for the test set reviews 
in all datasets. 
See the appendix for details of the data split and accuracies.

We use the BERT model fine-tuned on reviews to predict sentiment of sentences. %
Note that this prediction entails a distribution shift as sentences are shorter than full reviews used to fine-tune BERT models.
However, this is a common strategy for evaluating rationales in the interpretable machine learning literature and there exists evidence that transformer-based models provide strong performance despite the distribution shift in the form of reduced inputs \citep{deyoung2019eraser,hsu+etal:20,carton+rathore+tan:20}.\footnote{\citet{bastan-etal-2020-authors} investigates the reverse direction, i.e., from paragraph-level predictions to document-level predictions.}

\begin{figure}
  \centering
  \begin{subfigure}{0.45\textwidth}
  \centering
        \includegraphics[width=0.9\textwidth]{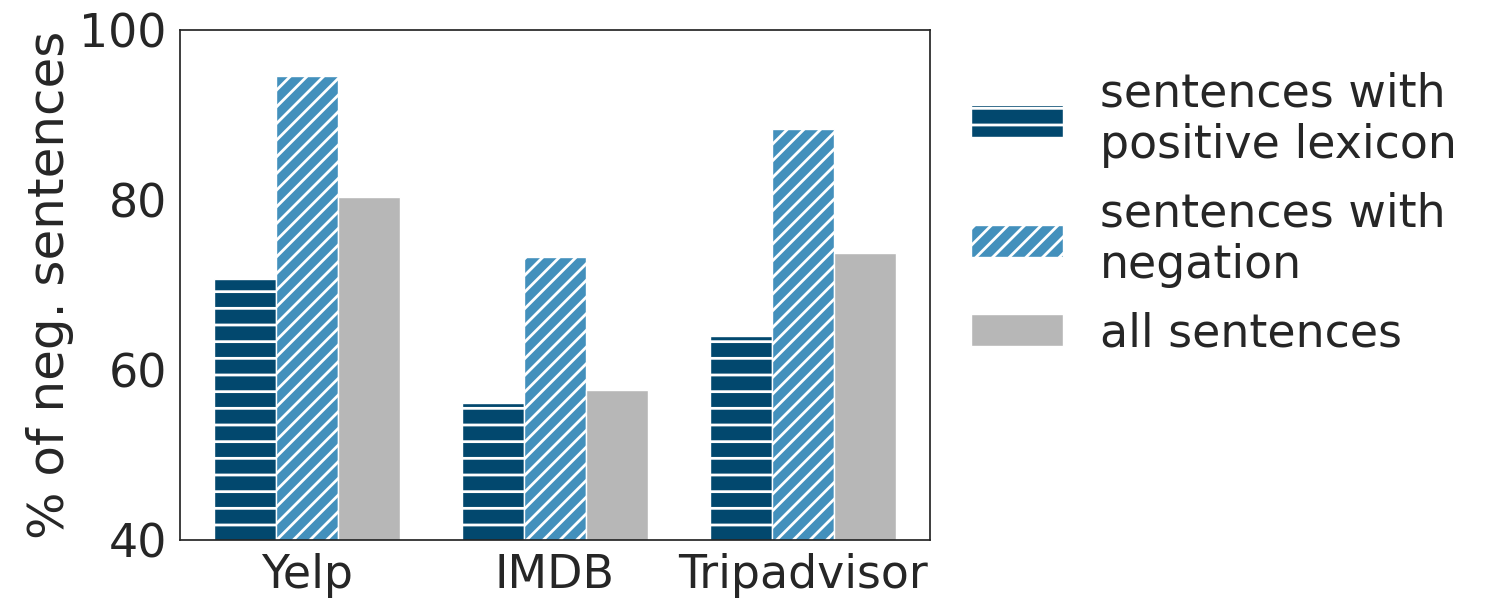}
      \caption{
    Fractions of negative sentences in negative reviews. 
      }
    \label{fig:bert_preds_negrev_non_amz}
  \end{subfigure}
  
  \begin{subfigure}{0.45\textwidth}
    \centering
        \includegraphics[width=0.9\textwidth]{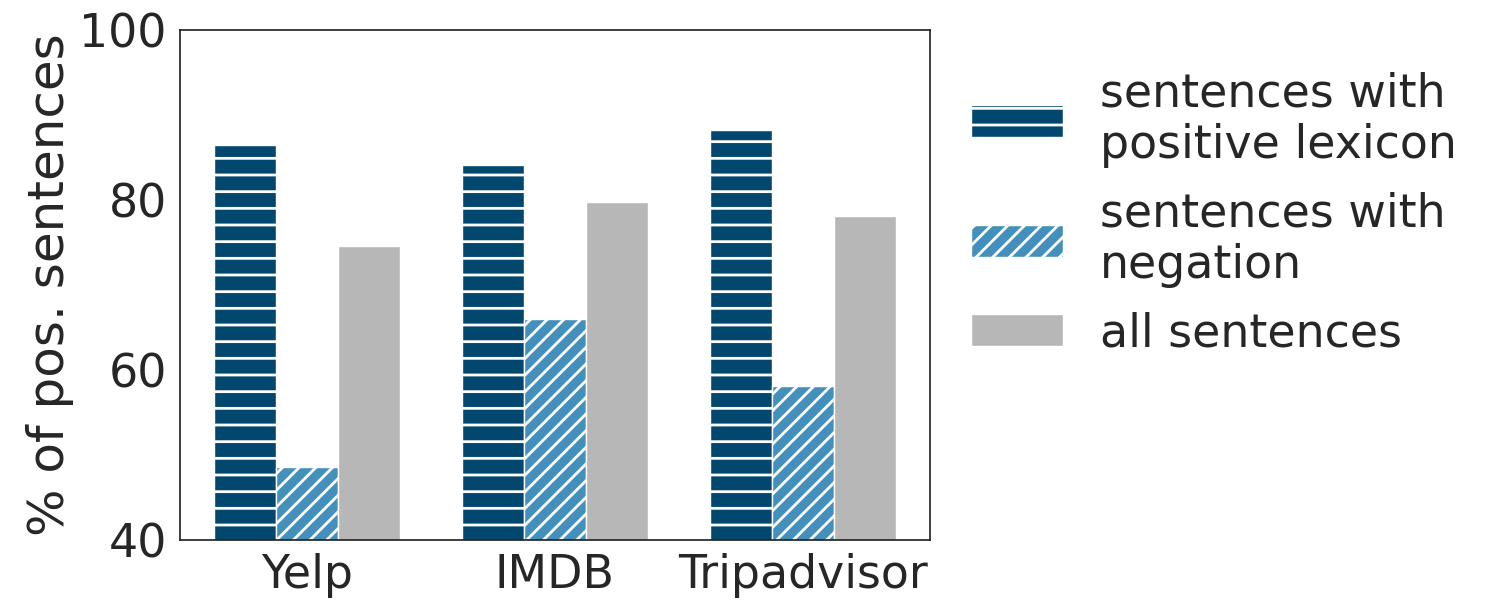}
      \caption{
    Fractions of positive sentences in positive reviews. 
      }
    \label{fig:bert_preds_posrev_non_amz}
  \end{subfigure}  
  \caption{Sentence-level prediction results based on fine-tuned BERT classifiers. In negative reviews, sentences with positive words tend to be negative, and sentences with negations are overwhelmingly negative. In comparison, sentences with negations are more balanced (44.7\% negative) in positive reviews.}
  \label{fig:bert_preds}
\end{figure}

\figref{fig:bert_preds_negrev_non_amz} shows 
that sentences with positive words in negative reviews are 
more 
likely to be negative than to be positive (65.2\% on average across all datasets; notably, IMDB is lower but still 
at 56.13\%,
above 50\%).%
\footnote{Similar trends hold if we adjust the estimates using TPR, TNR, FPR, and FNR. See the appendix.}
It suggests that the majority of positive words are negated in some way.
While the remaining minority of sentences with positive words are indeed positive and align with 
the tendency to report positive views,
our results highlight the important role of indirect expressions in explaining the 
positive words in negative reviews.

Furthermore, sentences with negation tend to be negative (88.6\%) based on our classifiers, confirming our hypothesis that most negations are used to express negative sentiments in negative reviews.
This is even higher than the average fraction of negative sentences 
(73.1\%)
among all sentences in negative reviews. 
In comparison, \figref{fig:bert_preds_posrev_non_amz} shows that positive words in positive reviews tend to reflect positive sentiments, indicating no common use of negation associated with positive words. Meanwhile, negations are not usually associated with negative sentiments in positive reviews (44.7\%), substantially lower than negations associated with negative sentiments in negative reviews (88.6\%).

\section{Conclusion}

In this paper, we 
investigate positivity bias in negative reviews and highlight the role of indirect expressions in understanding this phenomenon.
We show that negations followed by positive words are more prevalent than any other combination in negative reviews. 
Given that these indirect wordings  account for only 11.3\%
 of the occurrences of positive words in negative reviews, we further show that such sentences with positive words tend to be negative, based on sentiment classifiers.

While our findings support the prevalence of indirect expressions,
we do not take sentiment intensity into account.
In practice, ``not good'' provides a different meaning from ``not amazing''. 
We believe exploring the relationship between negation and semantic intensity is a promising 
direction. 
Our lexical-driven approaches are limited by the lexicons included in the dictionaries, which are typically evaluated independent of the context, so their sentiment may be different in the specific context.\footnote{One reviewer pointed out an interesting hypothesis: judges assume the nicest interpretation of a word out of context in the annotation process, as a result, the Pollyanna hypothesis may be an instrumentation bias.}
Similarly, our sentence-level prediction results are limited by
the distribution shift when applying BERT trained on documents to sentences.
It is reassuring that our high-level results hold across multiple datasets based on both lexical-driven approaches and sentence-level prediction.
As our study focuses on negative reviews in English, it is important to examine the generalizability of our results.
For instance, it is important to understand to what extent the observed positivity bias in general expressions is driven by such indirect expressions. 
Another natural extension is to investigate other languages.
Although our findings are limited to English reviews, we believe that they may be applicable to negative opinions in other languages, as Pollyanna hypotheses \cite{boucher1969pollyanna} has been validated across languages and cultures.
Finally, our work has implications for sentiment-related applications in NLP. 
The prevalence of indirect expressions in negative reviews underscores the importance of modeling and understanding negation in sentiment analysis and sentiment transfer \citep{ettinger2020bert}. 

In general, we believe that online reviews not only provide valuable data for teaching machines to recognize sentiments but also allow us to understand how humans express sentiments.
We hope that our work encourages future work to further investigate the framing choices when we express emotions and opinions, and 
their implications on NLP applications.

\section*{Acknowledgments}

We thank anonymous reviewers and the members of the Chicago Human+AI lab for their helpful comments.
This work was supported in part by an Amazon research award, a Salesforce research award, and NSF IIS-1941973.

\bibliography{refs}
\bibliographystyle{acl_natbib}

\clearpage
\appendix

\section{Vader Lexicons}
\label{appendix:vader}

\tableref{tab:negation_lexicons} shows the list of negation lexicons in Vader.

\begin{table}[h]
  \small
  \centering
  \begin{tabular}{ p{0.45\textwidth} } 
  \toprule
  aint, arent, cannot, cant, couldnt, darent, didnt, doesnt, ain't, aren't, can't, couldn't, daren't, didn't, doesn't, dont, hadnt, hasnt, havent, isnt, mightnt, mustnt, neither, don't, hadn't, hasn't, haven't, isn't, mightn't, mustn't, neednt, needn't, never, none, nope, nor, not, nothing, nowhere, oughtnt, shant, shouldnt, uhuh, wasnt, werent, oughtn't, shan't, shouldn't, uh-uh, wasn't, weren't, without, wont, wouldnt, won't, wouldn't, rarely, seldom, despite \\ 
  \bottomrule
  \end{tabular}
  \caption{Negation lexicons in Vader used for our negation analysis.}
  \label{tab:negation_lexicons}
\end{table}

\section{Samples from PeerRead}

\tableref{tab:peer_negation_samples} shows a list of 6 sentences with negation selected from random negative PeerRead reviews. Negations are mostly associated with positive words, both directly and indirectly.

\begin{table}[h]
  \small
  \centering
  \begin{tabular}{ p{0.45\textwidth} } 
  \toprule
  \textcolor{blue}{Please}  do \textit{not} make incredibly unscientific statements \textcolor{blue}{like} this one :``. \\ 
  \\
  I'm \textit{not} \textcolor{blue}{convinced}  about the \textcolor{blue}{value}  of having this artificial dataset. \\
  \\
  For example, at the end of sec 4.4, `` This result is \textit{not} \textcolor{blue}{surprising}, given that FOV-R contains additional information .... \\
  \\
  It is \textit{not} \textcolor{blue}{clear} whether the \textcolor{blue}{improvements} (if there is) of the ensemble disappear after data-augmentation. \\
  \\
  Empirical analysis is \textit{not} \textcolor{blue}{satisfactory}. \\
  \\
  But I'm \textit{not} \textcolor{blue}{sure} from reading the paper. \\
  \bottomrule
  \end{tabular}
  \caption{Sentences with negation sampled from negative reviews of PeerRead. Positive words are in blue and negative words are in red. Negations are in italics.}
  \label{tab:peer_negation_samples}
\end{table}

\begin{table*}[t]
  \centering
  \small
  \begin{tabular}{ccccc}
  \toprule
  Dataset & Training set & Validation set & Test set & Test accuracy (\%) \\
  \midrule
  Yelp & 45000 & 5000 & 10000 & 97.51 \\
  IMDB & 22500 & 2500 & 10000 & 94.38 \\
  Tripadvisor & 45000 & 5000 & 10000 & 96.66 \\
  Automotive & 45000 & 5000 & 10000 & 95.65 \\
  Cellphones and accessories & 45000 & 5000 & 10000 & 95.39\\
  Luxury beauty & 4195 & 467 & 3040 & 96.10\\
  Pet supplies & 45000 & 5000 & 10000 & 95.60\\
  Sports and outdoors & 45000 & 5000 & 10000 & 95.12\\
  \bottomrule
  \end{tabular}
  \caption{Dataset split and test accuracies of BERT fine-tuning. 
  For all datasets except IMDB, Luxury Beauty, we use 45K samples as training set, 5K as validation set, and 10K as test set, randomly sampled from the entire dataset. In the case of IMDB, we use 22.5K samples for training and 2.5K samples for validation, randomly sampled from the provided training set of size 25K. We then use 10K samples randomly sampled from the provided test set of size 25K for testing purposes. In the case of Amazon Luxury Beauty, we use a balanced set of 4195 samples for training and 467 samples for validation. We then use 3K samples (imbalanced) randomly sampled from the dataset for testing. All these random samplings were done without replacement.
  }
  \label{tab:dataset_split_and_accs_sent_analysis}
\end{table*}

\section{Additional Plots}

\begin{figure}[h!]
\centering
\begin{subfigure}[t]{0.4\textwidth}
  \centering
    \includegraphics[width=0.95\textwidth]{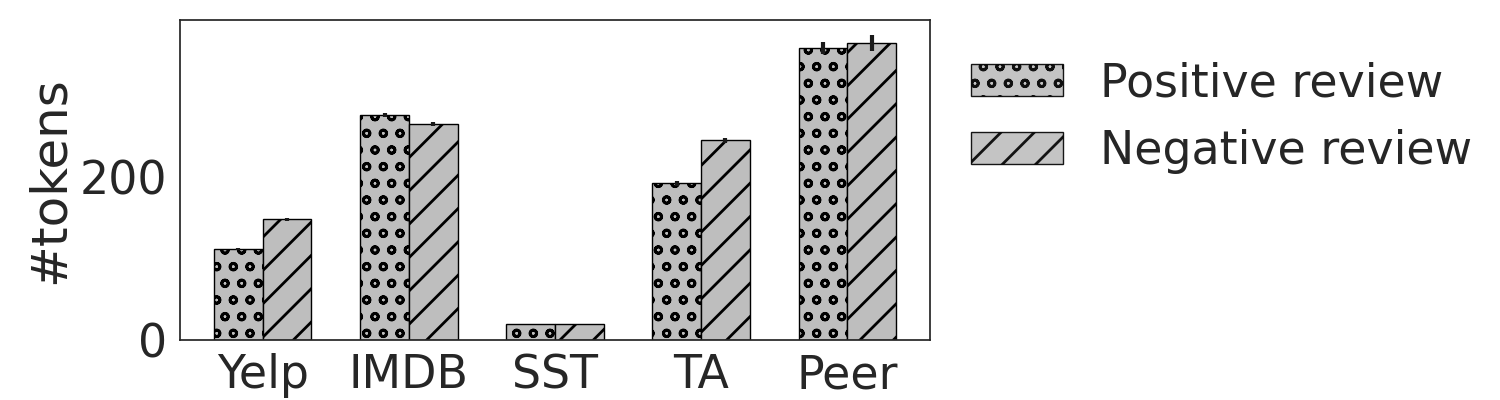}
  \caption{SST, Yelp, IMDB, and Tripadvisor (non-Amazon datasets).
  }
  \label{fig:review_length_cmp_non_amz}
\end{subfigure}
\begin{subfigure}[t]{0.4\textwidth}
  \centering
    \includegraphics[width=0.95\textwidth]{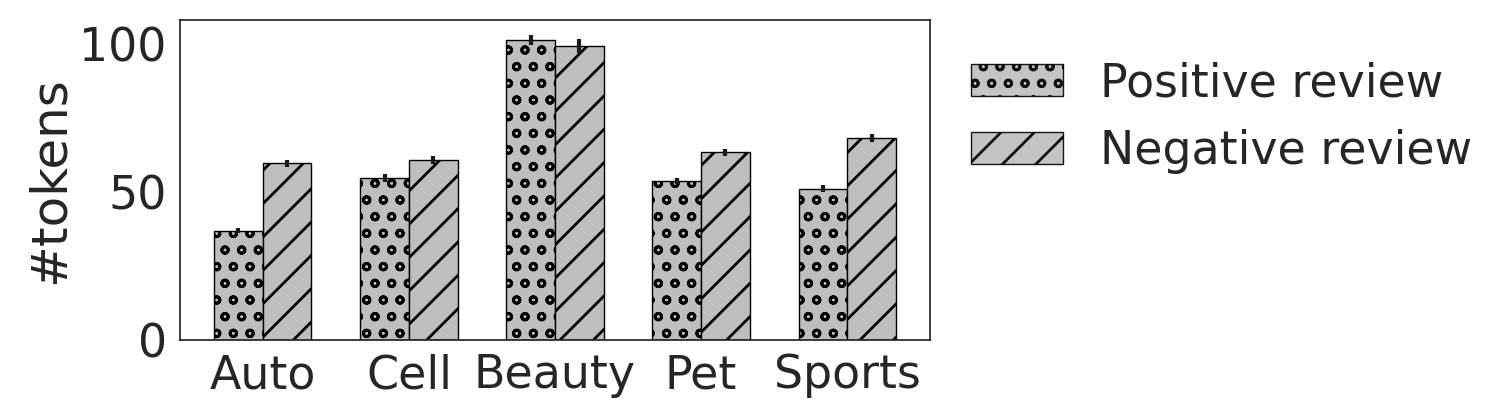}
  \caption{Amazon datasets.
  }
  \label{fig:review_length_cmp_amz}
\end{subfigure}
\caption{Review-level length distribution. This shows the length comparisons of positive and negative reviews of different datasets. The values represent the average number of tokens present in each review. Negative reviews are longer than positive reviews in all datasets except IMDB and Amazon Luxury Beauty.
}
\label{fig:revew_length_cmp_all}
\end{figure}

\begin{figure}[h!]
\centering
  \includegraphics[width=0.4\textwidth]{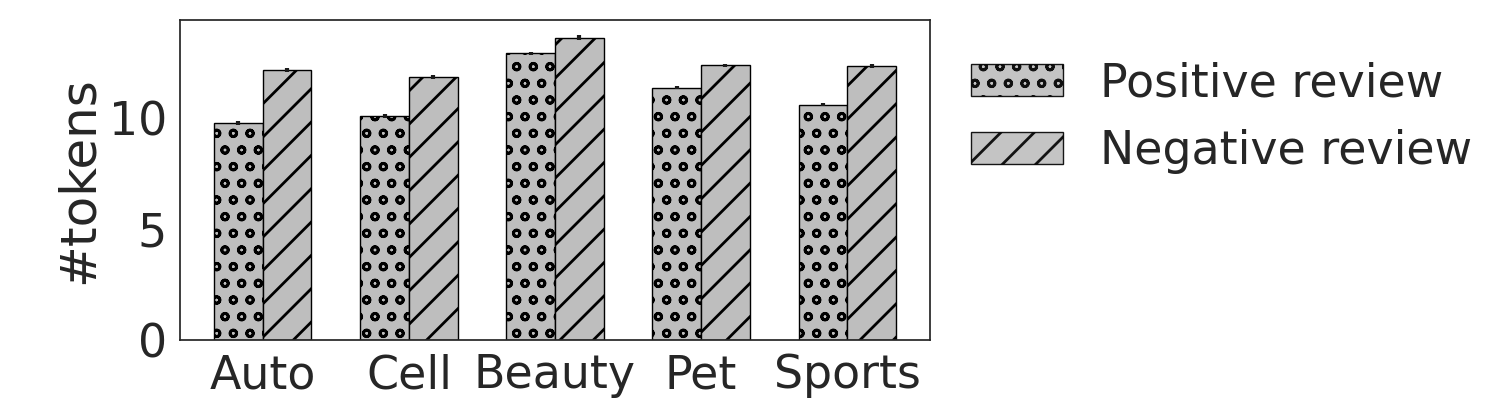}
  \caption{Sentence-level length distribution of Amazon datasets. 
  }
\label{fig:sentence_lenght_amz}
\end{figure}

\para{Length distribution}. See \figref{fig:revew_length_cmp_all} for review-level length and \figref{fig:sentence_lenght_amz} for sentence-level length distribution for Amazon reviews. 

\begin{figure}[h!]
\centering
\begin{subfigure}[t]{0.48\textwidth}
  \centering
\includegraphics[width=0.95\textwidth]{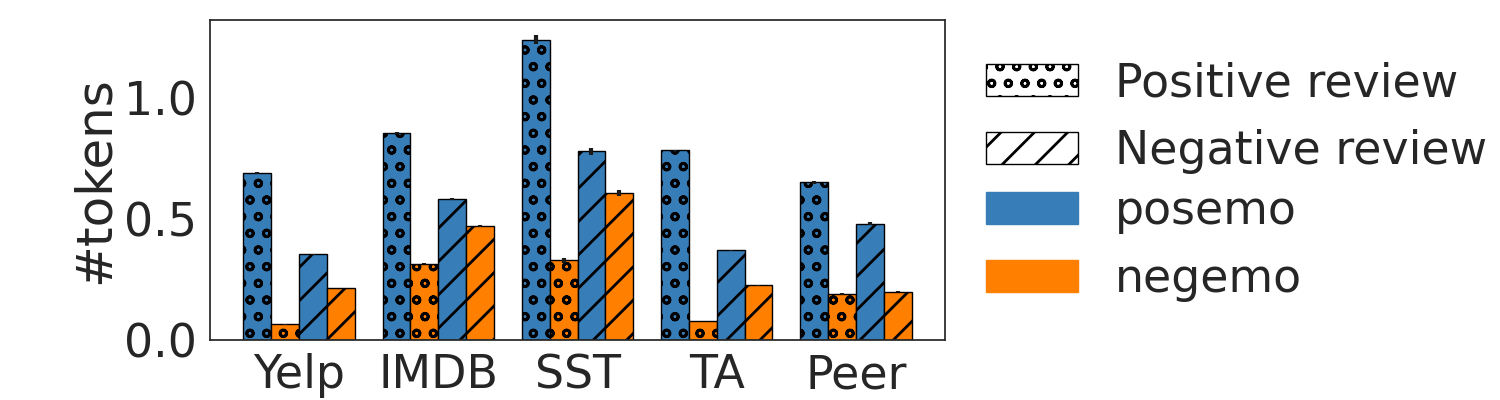}
  \caption{
  Non-Amazon datasets.
  }
  \label{fig:liwc_pos_neg_non_amz}
\end{subfigure}
\begin{subfigure}[t]{.48\textwidth}
  \centering
  \includegraphics[width=0.95\textwidth]{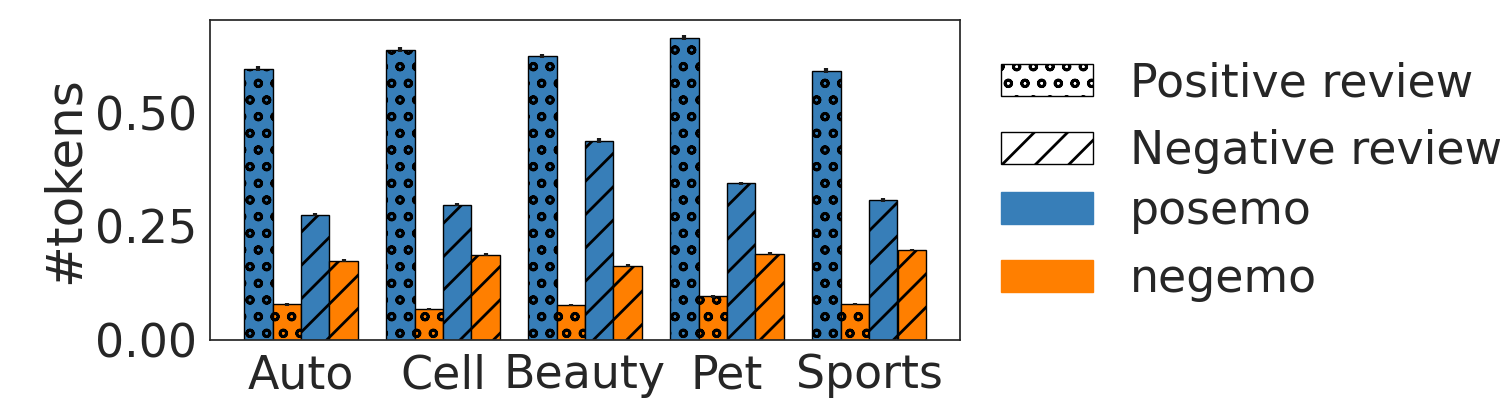}
  \caption{Amazon datasets.}
  \label{fig:liwc_pos_neg_amz}
\end{subfigure}
\caption{Lexicon distributions based on LIWC. \figref{fig:liwc_pos_neg_non_amz} and \figref{fig:liwc_pos_neg_amz} shows the lexicon distribution of reviews using \textit{posemo} and \textit{negemo} LIWC categories. In all datasets, negative reviews have fewer positive emotions than positive reviews. They also have more positive words than negative words. This trend is similar to that obtained using Vader lexicons in case of non-Amazon reviews. 
}
\label{fig:liwc_pos_neg_all}
\end{figure}

\begin{figure}[h!]
\centering
\includegraphics[width=0.45\textwidth]{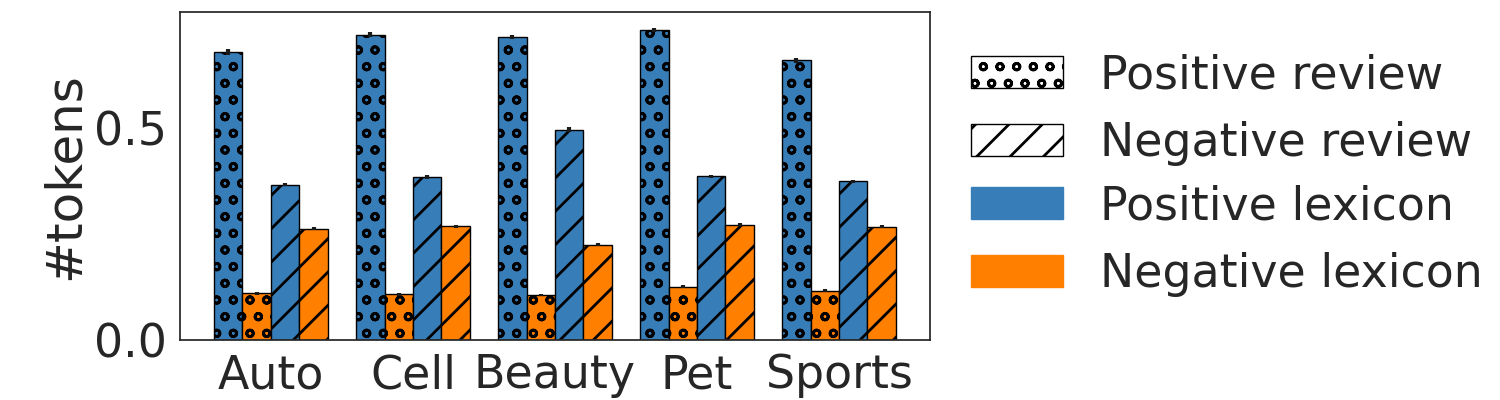}
\caption{Lexicon distribution of Amazon datasets using Vader. Negative reviews have more positive words than negative words, similar to the trend in SST, Yelp, IMDB, Tripadvisor, and PeerRead.
}
\label{fig:vader_pos_neg_amz}
\end{figure}

\para{Lexicon distribution}. \figref{fig:liwc_pos_neg_all} shows the sentiment lexicon distribution of all reviews using LIWC. \figref{fig:vader_pos_neg_amz} shows the lexicon distribution of Amazon reviews using Vader. 

\begin{figure}[h!]
\centering
\begin{subfigure}[t]{0.4\textwidth}
  \centering
  \includegraphics[width=0.95\textwidth]{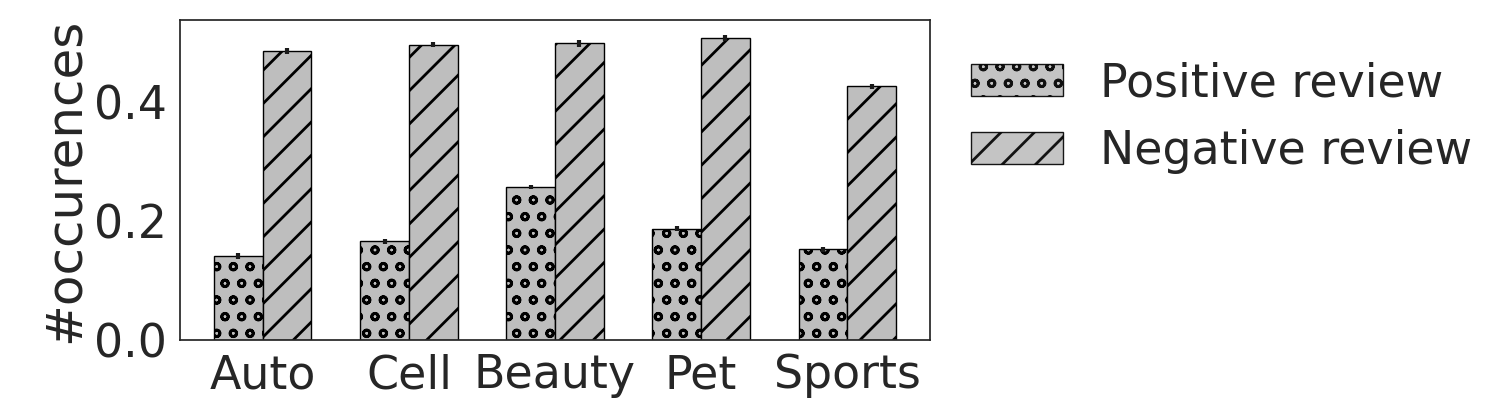}
  \caption{Overall negation.
  }
  \label{fig:overall_negation_amz}
\end{subfigure}
\begin{subfigure}[t]{.4\textwidth}
  \centering
  \includegraphics[width=0.95\textwidth]{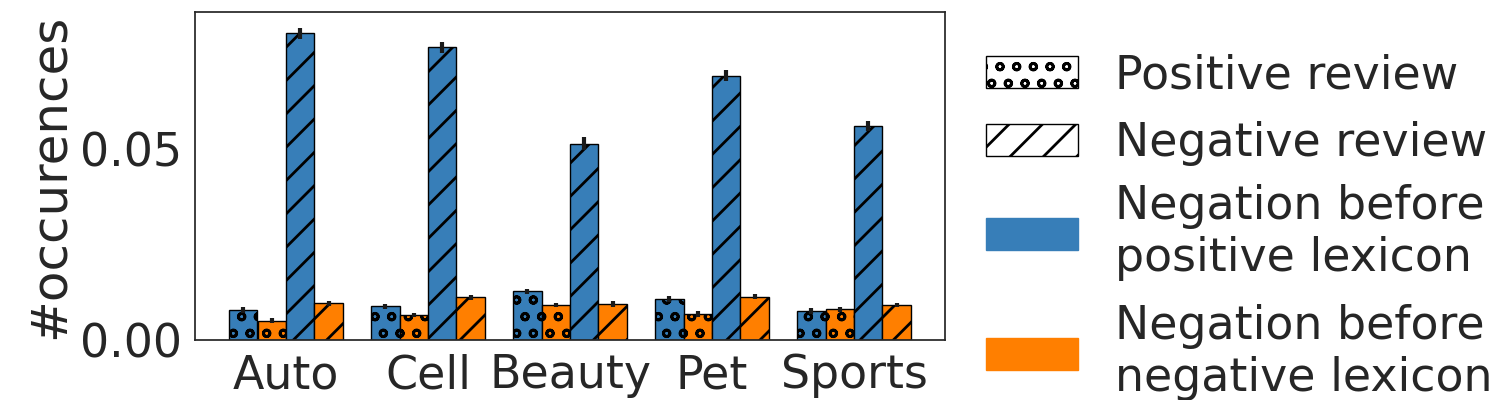}
  \caption{Negation before positive and negative lexicons.
  }
  \label{fig:vader_sentiment_negation_amz}
\end{subfigure}
\caption{Negation distribution of Amazon datasets using Vader lexicons. Negative reviews use more negation words compared to positive reviews. Negative reviews have substantially more negation words associated with positive words than negative words.
}
\label{fig:vader_negation_amz}
\end{figure}

\begin{figure}[h!]
\centering
\begin{subfigure}[t]{.48\textwidth}
  \centering
  \includegraphics[width=0.95\textwidth]{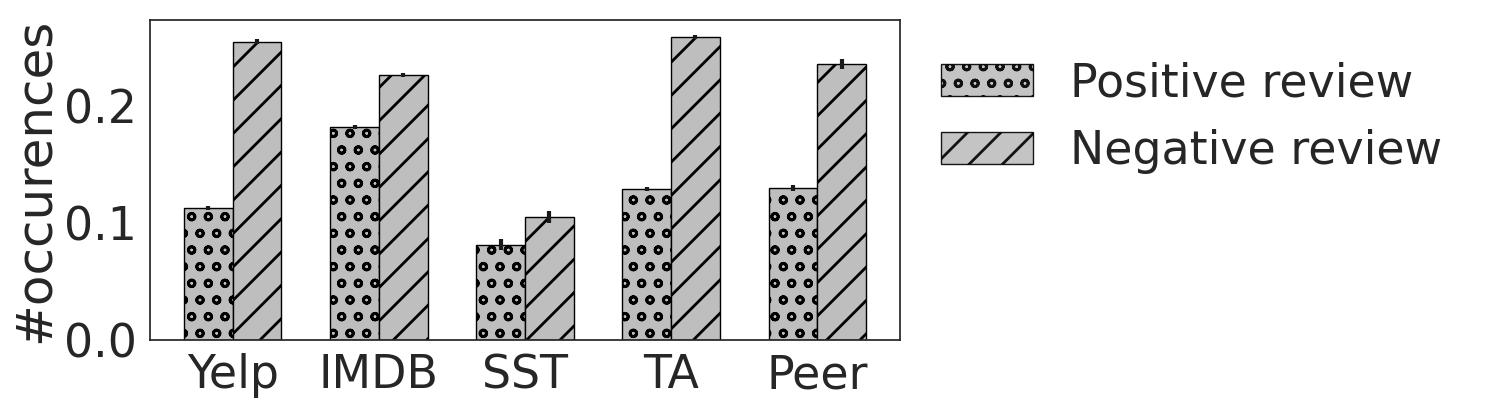}
  \caption{Overall negation.
  }
  \label{fig:dp_overall_negation_non_amz}
\end{subfigure}
\begin{subfigure}[t]{.48\textwidth}
  \centering
  \includegraphics[width=0.95\textwidth]{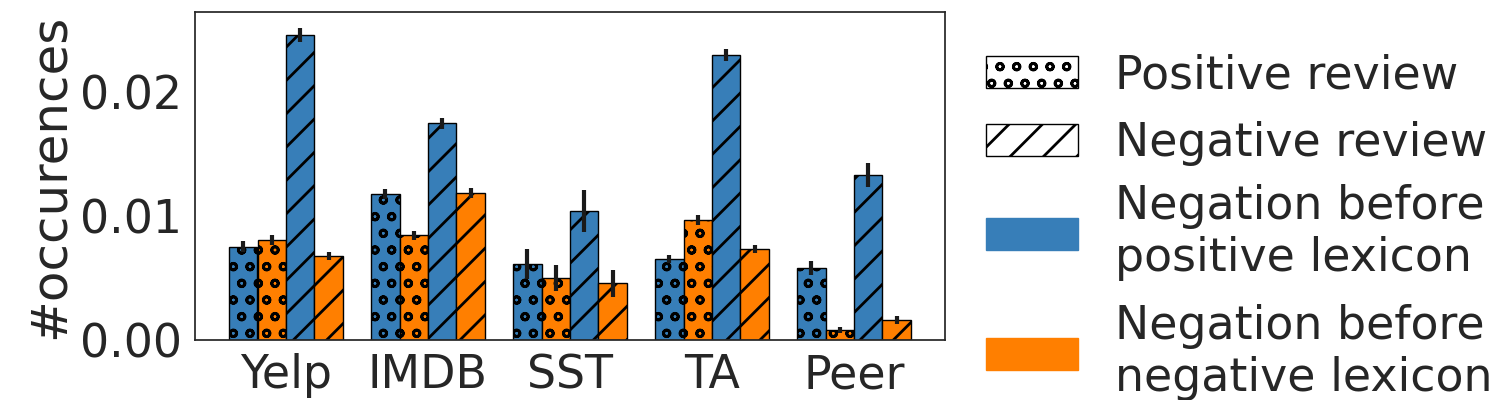}
  \caption{Negation before positive and negative lexicons.
  }
  \label{fig:dp_sentiment_negation_non_amz}
\end{subfigure}
\caption{Negation distribution using dependency parsing - non-Amazon datasets. In all non-Amazon datasets, negative reviews use more negation words than positive reviews. This observation is inline with the negation results obtained using Vader lexicons. Dependency parsing is used
to extract negations from reviews, and to identify words associated with a negation word. }
\label{fig:dp_negation_non_amz}
\end{figure}

\begin{figure}[t]
\centering
\begin{subfigure}[t]{0.4\textwidth}
  \centering
  \includegraphics[width=0.95\textwidth]{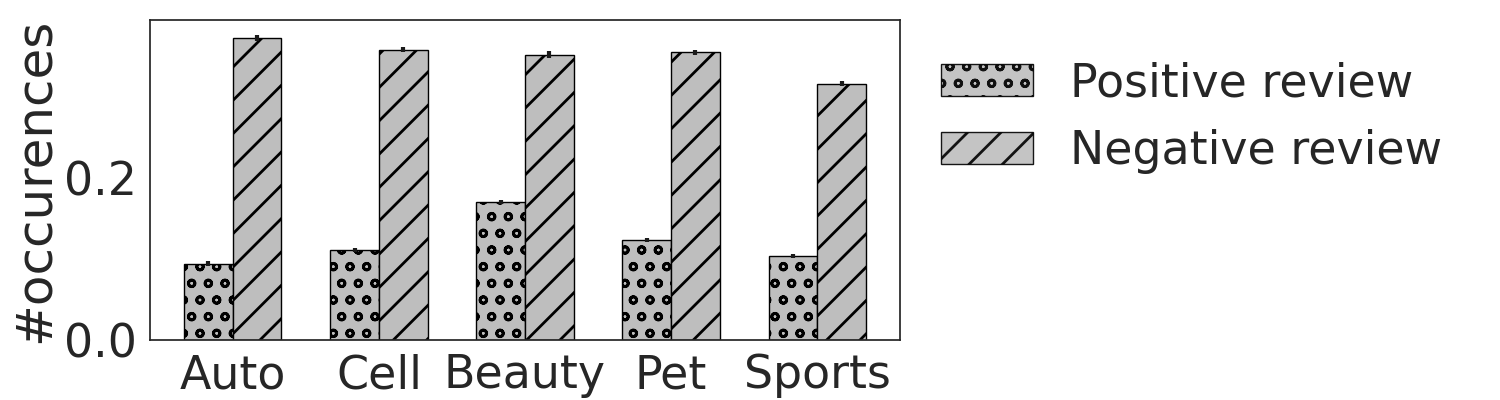}
  \caption{Overall negation.
  }
  \label{fig:dp_overall_negation_amz}
\end{subfigure}
\begin{subfigure}[t]{0.4\textwidth}
  \centering
  \includegraphics[width=0.95\textwidth]{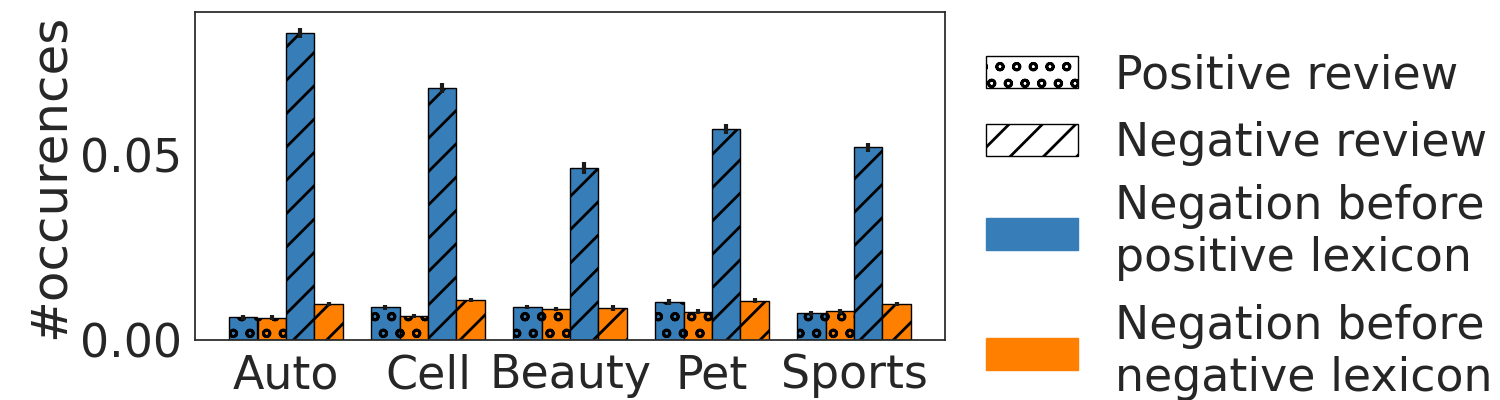}
  \caption{Negation before positive and negative lexicons.
  }
  \label{fig:dp_sentiment_negation_amz}
\end{subfigure}
\caption{Negation distribution using dependency parsing - Amazon datasets. \figref{fig:dp_overall_negation_amz} shows that negative reviews have substantially more negation words than positive words. \figref{fig:dp_sentiment_negation_amz} shows the negation distribution associated with positive and negative words. This corresponds to 
about 16.68\% of all negation words used in the positive and negative reviews based on our dictionary. 
Negative reviews also have substantially more negations before positive words, compared to other combinations.
}
\label{fig:dp_negation_amz}
\end{figure}

\para{Negation distribution}. 
See \figref{fig:vader_negation_amz} and \figref{fig:dp_negation_amz} for the negation distribution of Amazon reviews using Vader and dependency parsing respectively. \figref{fig:dp_negation_non_amz} shows the negation distribution found using dependency parsing for non-Amazon reviews.

\begin{figure}
\centering
  \includegraphics[width=0.4\textwidth]{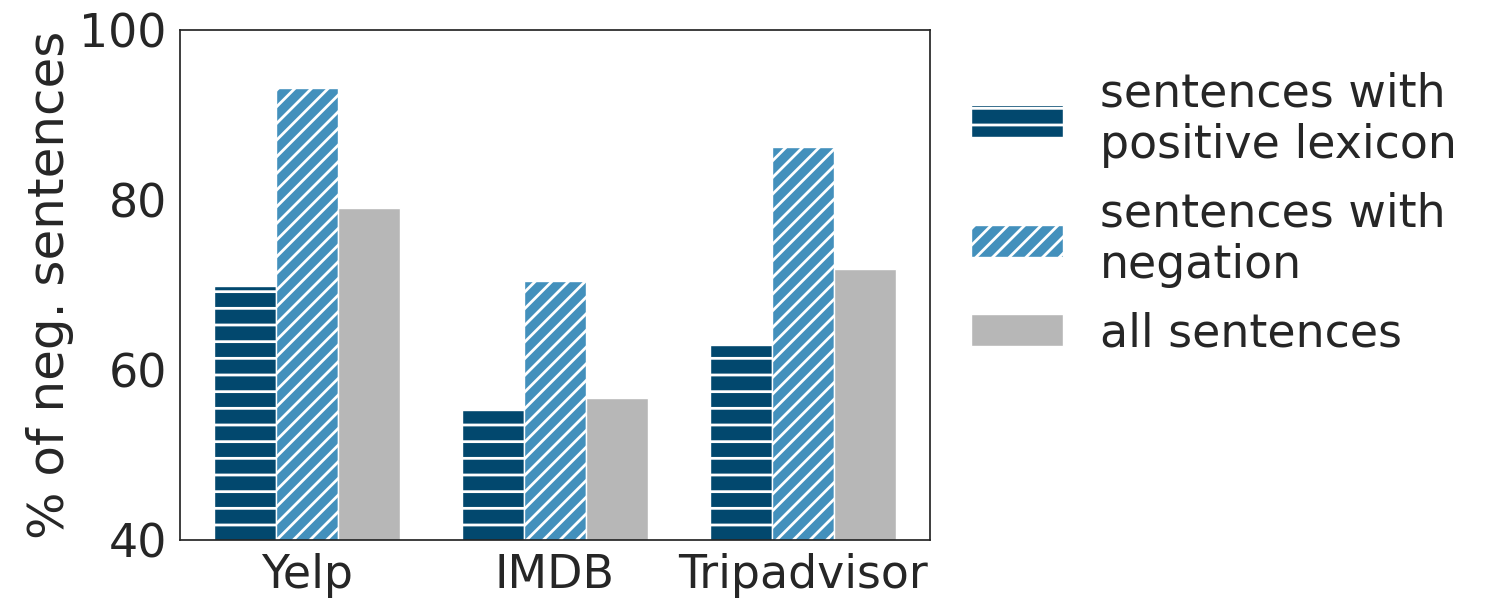}
\caption{Fractions of negative sentences in negative reviews of Yelp, IMDB, and Tripadvisor.
These fractions are corrected using TPR, TNR, FPR, and FNR. It can be seen that higher proportion of negative reviews with negation are classified as negative by our BERT model. 
This shows that negations in negative reviews are mostly used to express negative opinions. This observation holds for other datasets also.} 
  \label{fig:bert_preds_negrev_corr_non_amz}
\end{figure}

\begin{figure}
\centering
\begin{subfigure}[t]{0.4\textwidth}
  \centering

    \includegraphics[width=0.95\textwidth]{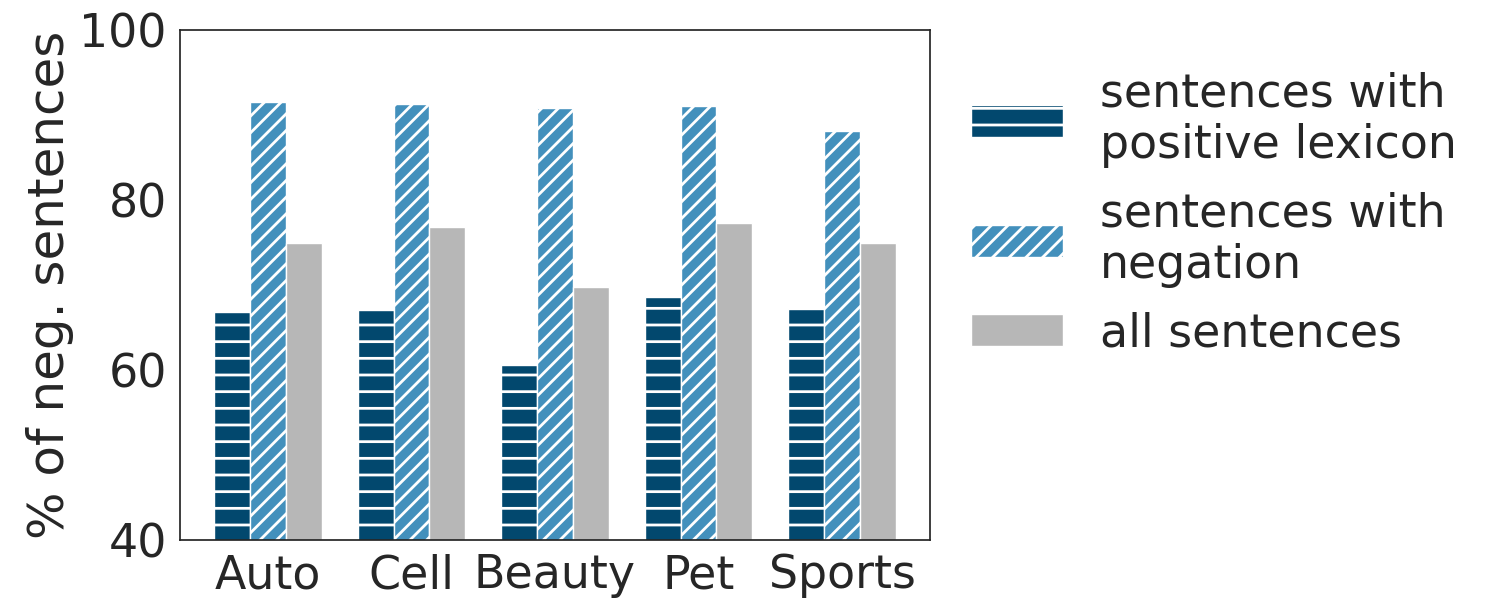}
\caption{ Fractions based on accuracy.}
\label{fig:bert_preds_negrev_amz}
\end{subfigure}

\begin{subfigure}[t]{0.4\textwidth}
\centering
    \includegraphics[width=0.95\textwidth]{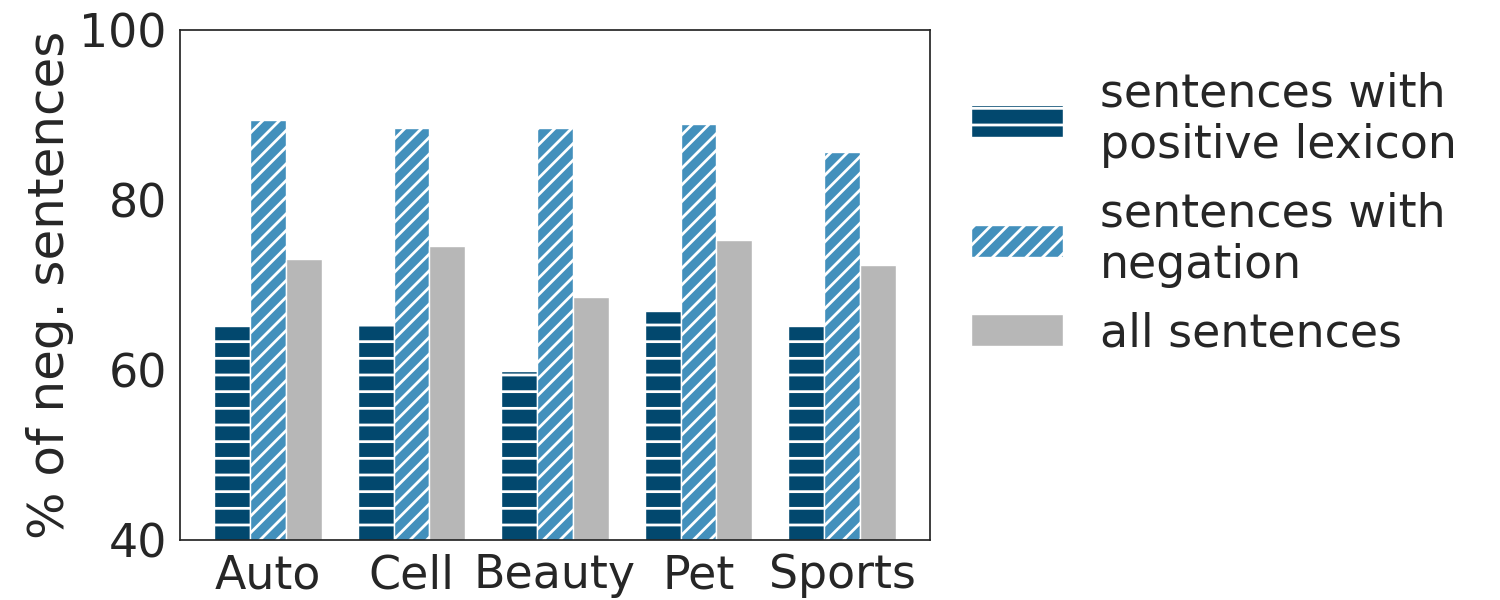}
\caption{Fractions based on TPR, TNR, FPR, and FNR.}
\label{fig:bert_preds_negrev_corr_amz}
\end{subfigure}
\caption{Fractions of negative sentences in negative Amazon reviews based on fine-tuned BERT classifiers. The distribution confirms our hypothesis that most negations are used to express negative sentiments.}
\label{fig:bert_preds_negrev_all_amz}
\end{figure}

\begin{figure}
\centering
\begin{subfigure}[t]{0.4\textwidth}
  \centering
  \includegraphics[width=0.95\textwidth]{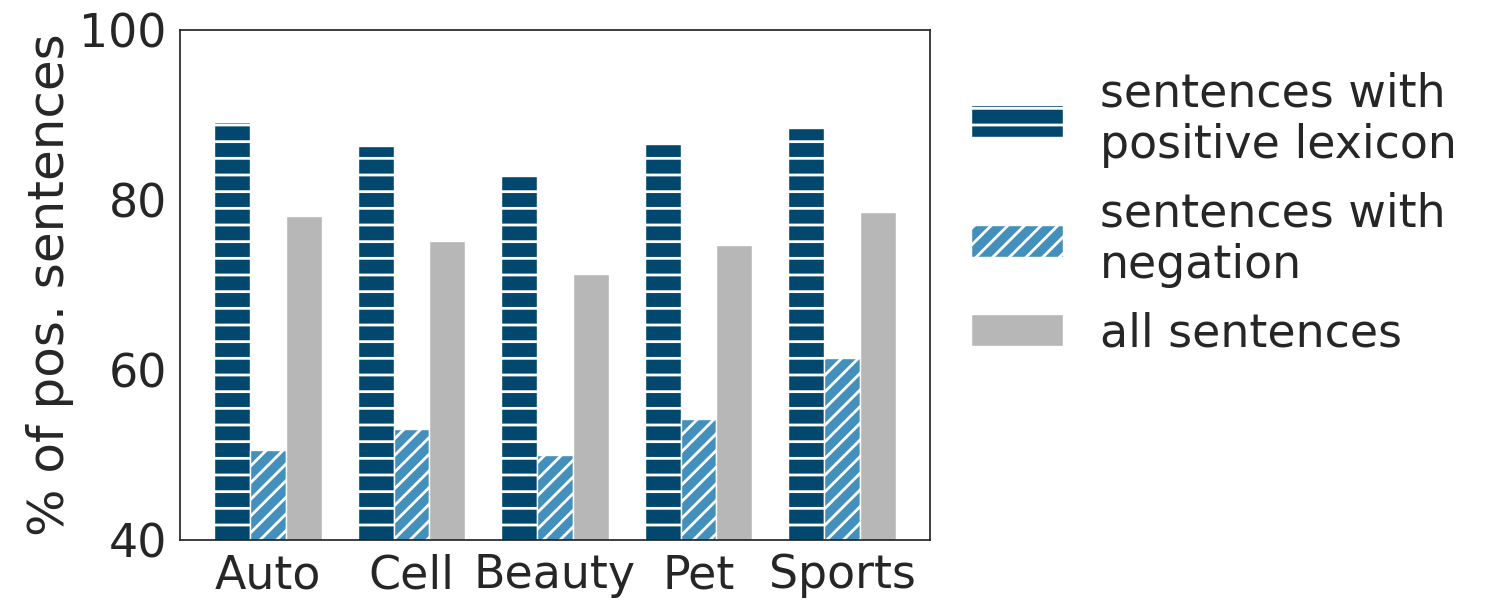}
  \caption{Fractions based on accuracy.}
  \label{fig:bert_preds_posrev_amz}
\end{subfigure}
\begin{subfigure}[t]{0.4\textwidth}
  \centering
  \includegraphics[width=0.95\textwidth]{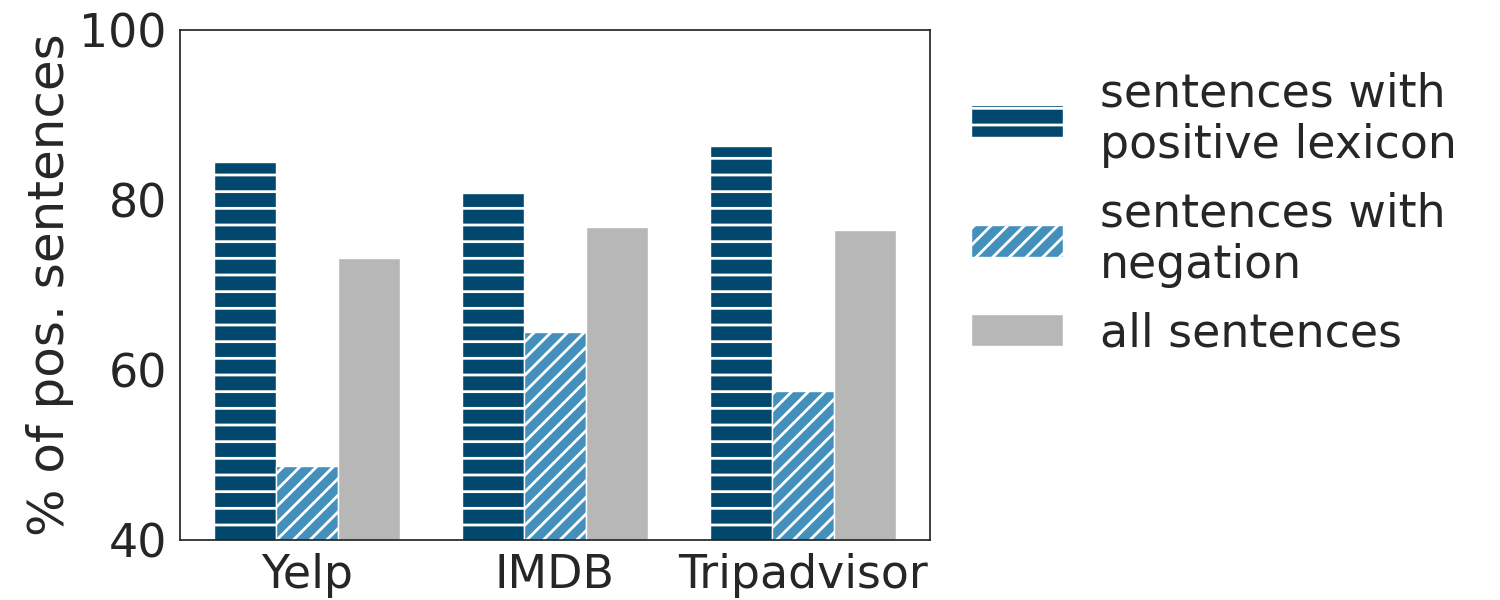}
  \caption{Fractions based on TPR, TNR, FPR, and FNR.}
  \label{fig:bert_preds_posrev_corr_non_amz}
\end{subfigure}
\begin{subfigure}[t]{0.4\textwidth}
  \centering
  \includegraphics[width=0.95\textwidth]{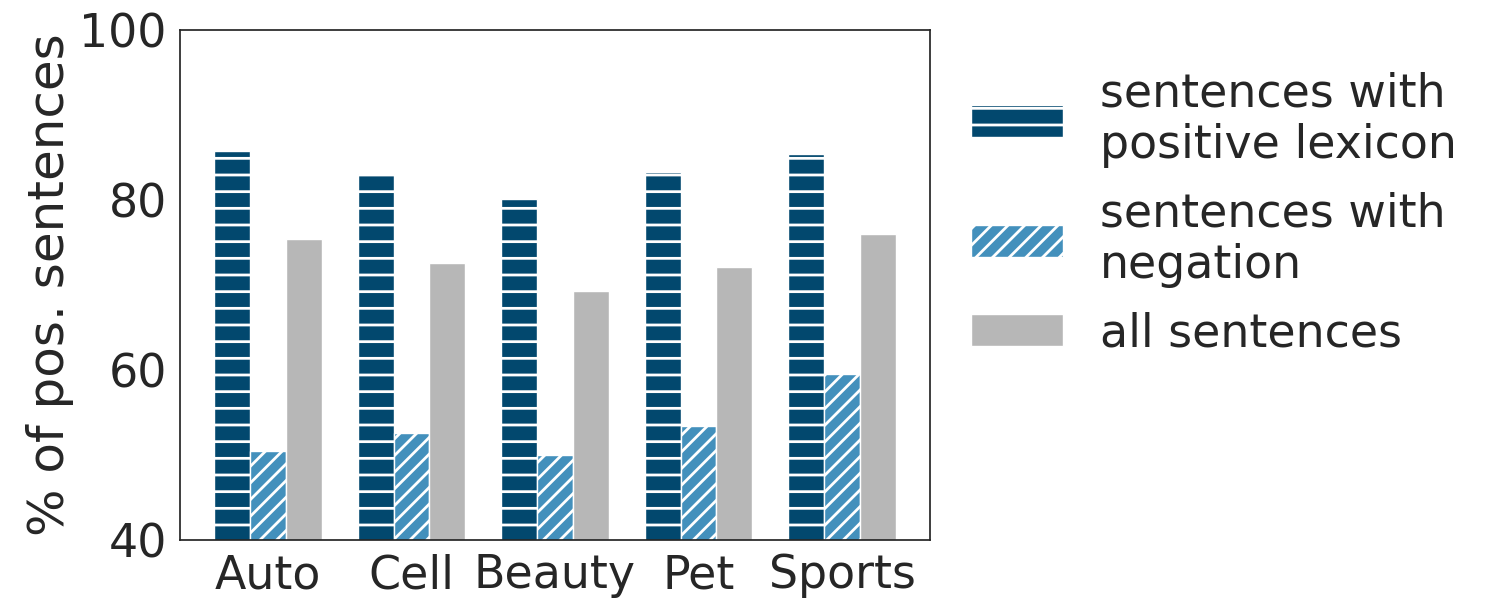}
  \caption{Fractions based on TPR, TNR, FPR, and FNR.}
  \label{fig:bert_preds_posrev_corr_amz}
\end{subfigure}
\caption{Fractions of positive sentences in positive reviews. We can see that negations in positive reviews are more balanced with positive and negative sentences when compared to negative reviews. 
Also, sentences with positive lexicons are mostly positive (86.5\%). 
There are very few negative sentences with positive lexicons. This holds for all datasets. 
} 
\label{fig:bert_preds_posrev_all}
\end{figure}

\para{Sentiment predictions}. See \figref{fig:bert_preds_negrev_non_amz} and \figref{fig:bert_preds_negrev_corr_non_amz} for the fractions of negative sentences in negative non-Amazon reviews measured by the BERT model.  
See \figref{fig:bert_preds_negrev_all_amz} for fractions of negative sentences in negative reviews of Amazon. \figref{fig:bert_preds_posrev_all} shows the fractions of positive sentences in positive reviews. Some of the fractions in our results are computed are based on the TPR, TNR, FPR, and FNR of the BERT model. We used test set reviews of the datasets to compute these metrics as they give more accurate estimate of percentage of positive and negative sentences in reviews.
All BERT classifiers that we used for predicting the sentiment of sentences are fine-tuned using the reviews of corresponding datasets. \tableref{tab:dataset_split_and_accs_sent_analysis} shows the dataset split and test accuracies of BERT fine-tuning.

\para{Hyperparameter tuning}. We did hyperparameter tuning by varying number of epochs, batch size, and learning rate. We fine-tuned BERT for 4 epochs with batch sizes of  2, 4 and 8, with a learning rates of 1e-5 and 2e-5. Based on validation accuracies, the model trained for 2 epochs, with a batch size of 8 and learning rate of 2e-5 turned out to be the best performing model for most of the datasets. 

\begin{figure*}
\centering
\begin{subfigure}[t]{.48\textwidth}
  \centering
  \includegraphics[width=0.95\textwidth]{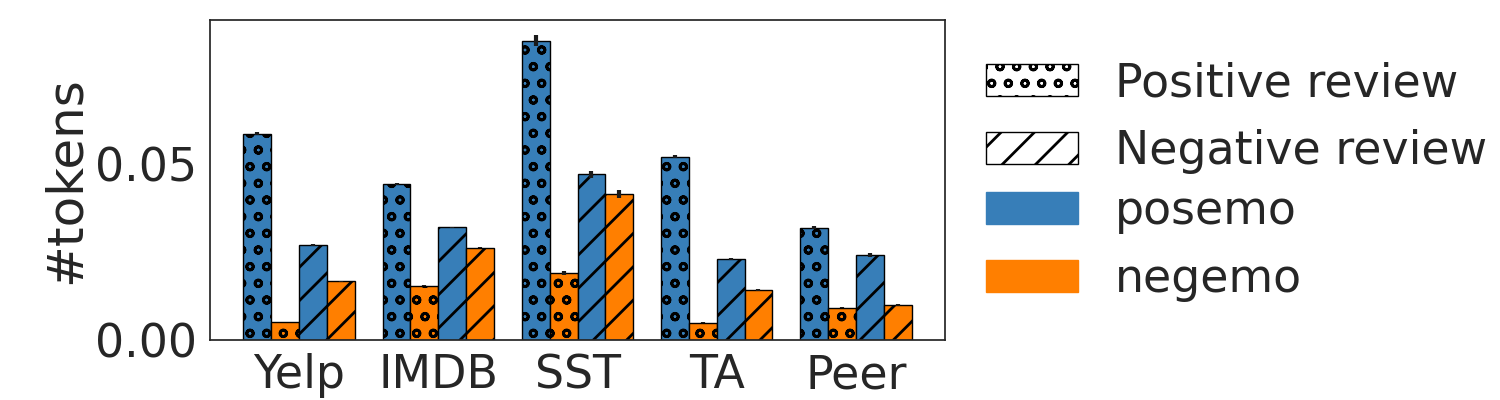}
  \caption{LIWC distribution of non-Amazon datasets.}
  \label{fig:liwc_pos_neg_dist_w_lvl_non_amz}
\end{subfigure}
\begin{subfigure}[t]{.48\textwidth}
  \centering
  \includegraphics[width=0.95\textwidth]{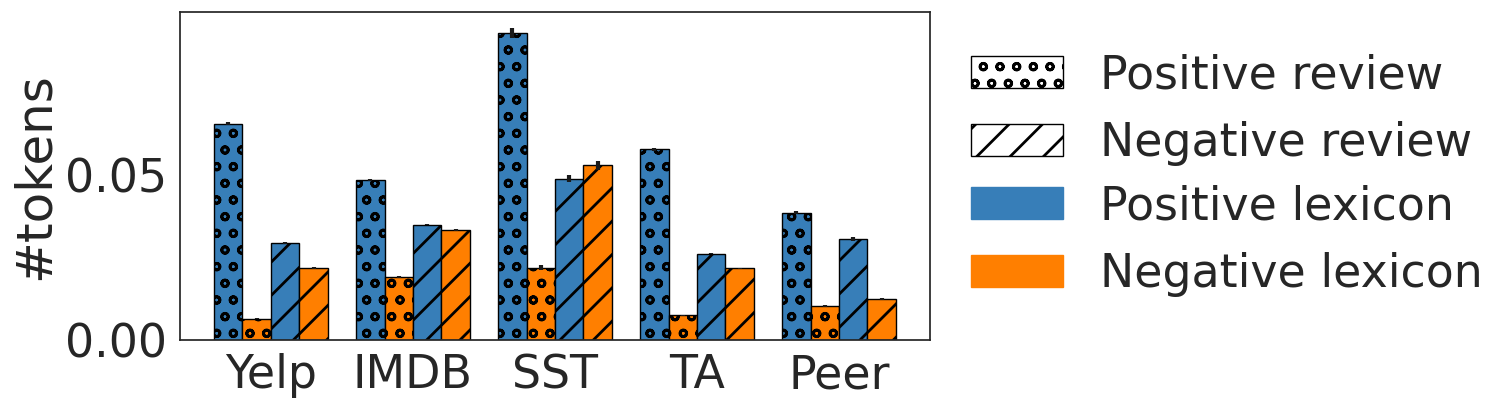}
  \caption{Vader distribution of non-Amazon datasets.}
  \label{fig:vader_pos_neg_dist_w_lvl_non_amz}
\end{subfigure}
\begin{subfigure}[t]{.48\textwidth}
  \centering
  \includegraphics[width=0.95\textwidth]{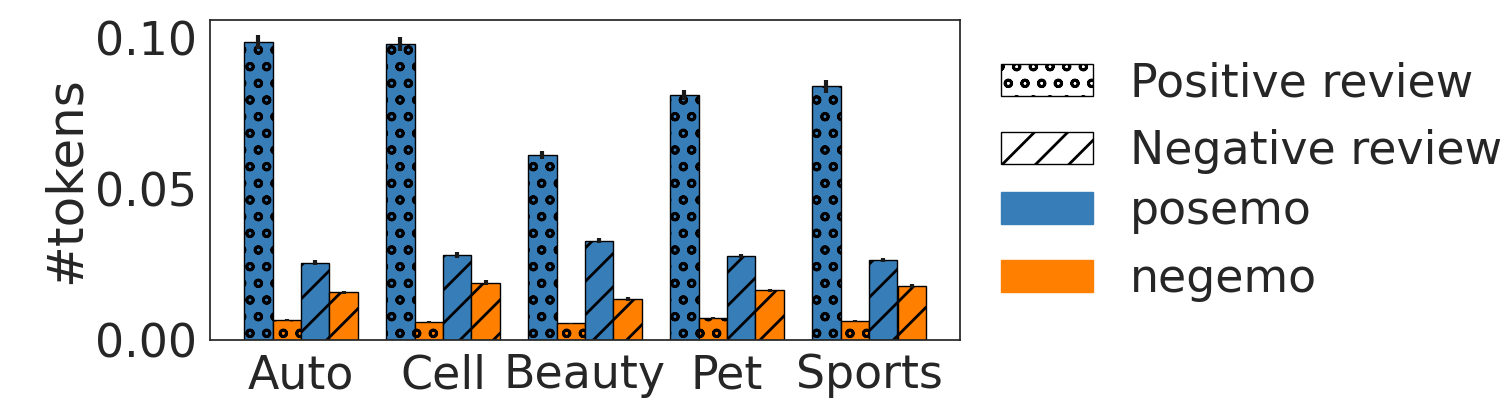}
  \caption{LIWC distribution of Amazon datasets.
  }
  \label{fig:liwc_pos_neg_dist_w_lvl_amz}
\end{subfigure}
\begin{subfigure}[t]{.48\textwidth}
  \centering
  \includegraphics[width=0.95\textwidth]{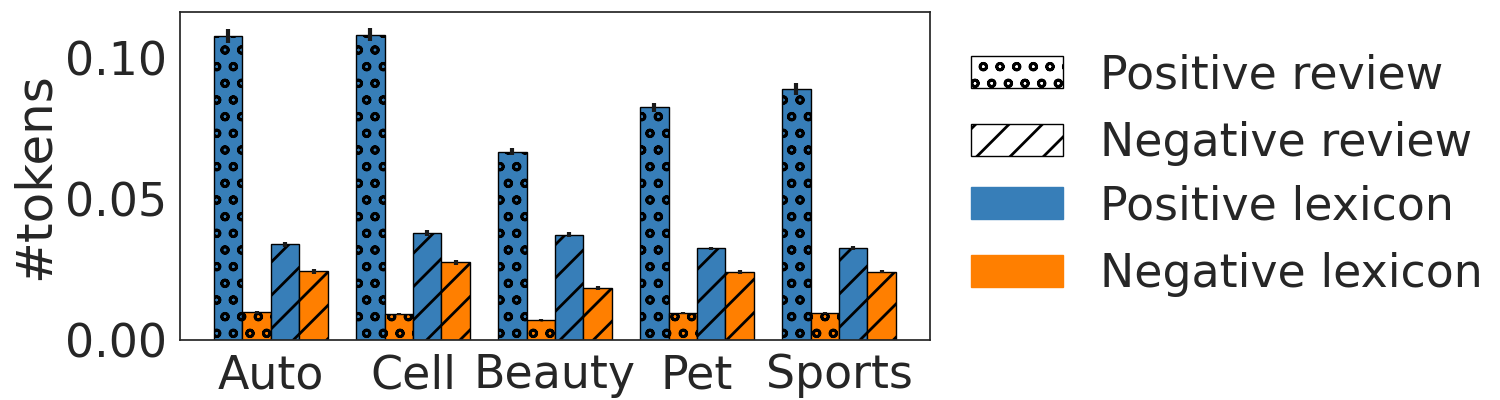}
  \caption{Vader distribution of Amazon datasets.
  }
  \label{fig:vader_pos_neg_dist_w_lvl_amz}
\end{subfigure}
\caption{Word-level lexicon distribution. At the word-level, positive reviews have more positive words than negative reviews. However, negative reviews contain more positive words than negative words (except SST with Vader).  The trend that we observe in the sentence-level results can be seen here as well.
}
\label{fig:lexicon_dist_w_lvl_all}
\end{figure*}

\begin{figure*}
\centering
\begin{subfigure}[t]{.48\textwidth}
  \centering
  \includegraphics[width=0.95\textwidth]{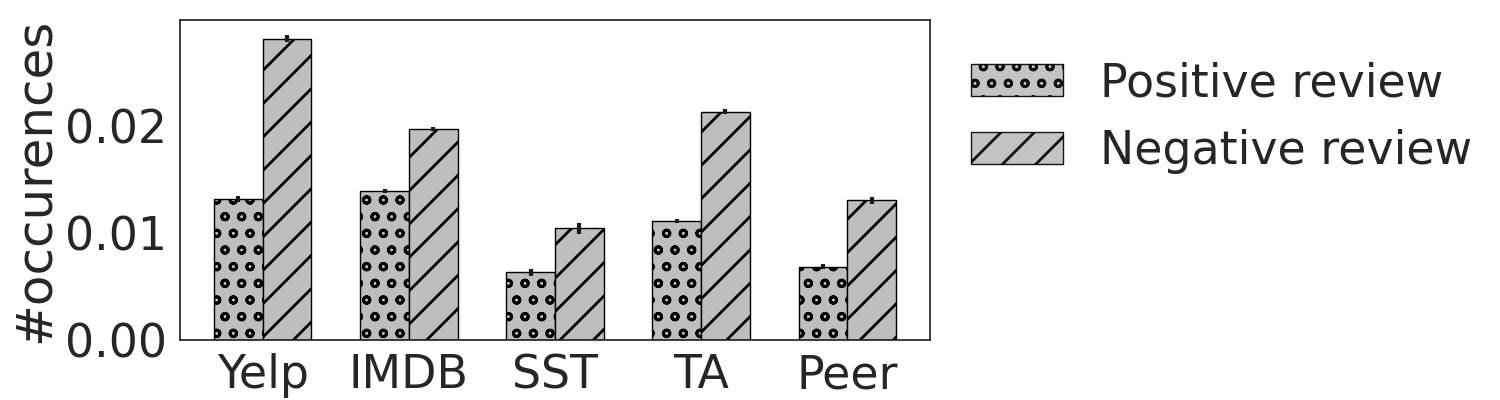}
  \caption{Non-Amazon datasets.
  }
  \label{fig:vader_overall_negation_w_lvl_non_amz}
\end{subfigure}
\begin{subfigure}[t]{.48\textwidth}
  \centering
  \includegraphics[width=0.95\textwidth]{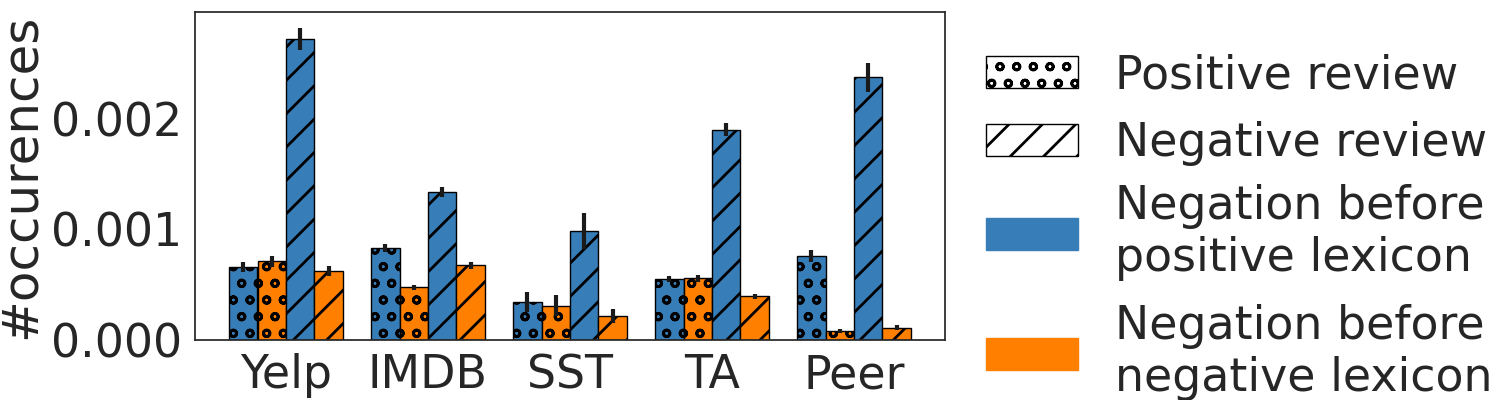}
  \caption{Non-Amazon datasets.
  }
  \label{fig:vader_sentiment_negation_w_lvl_non_amz}
\end{subfigure}
\begin{subfigure}[t]{.48\textwidth}
  \centering
  \includegraphics[width=0.95\textwidth]{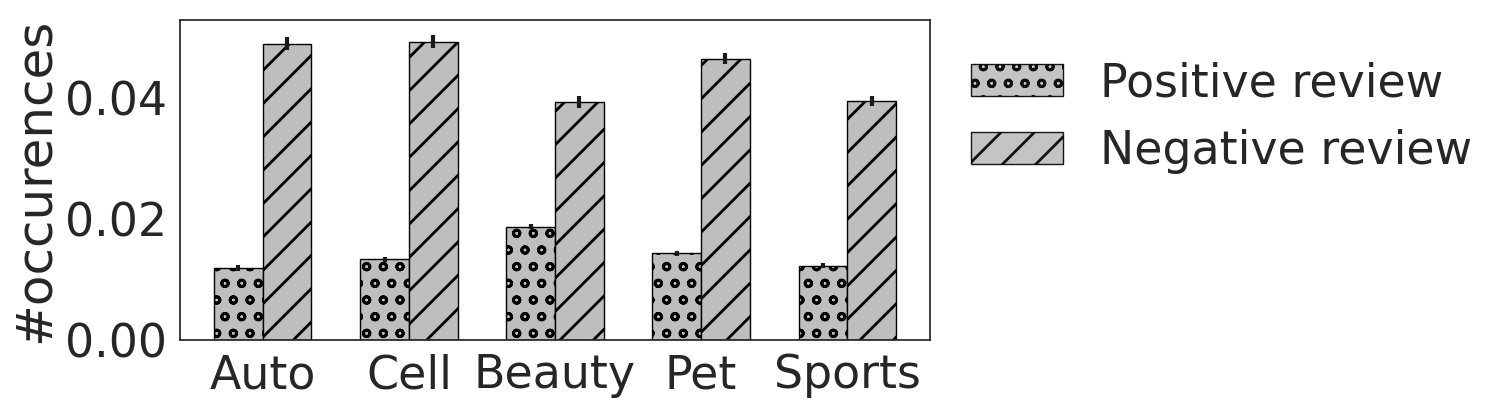}
  \caption{Amazon datasets.}
  \label{fig:vader_overall_negation_w_lvl_amz}
\end{subfigure}
\begin{subfigure}[t]{.48\textwidth}
  \centering
  \includegraphics[width=0.95\textwidth]{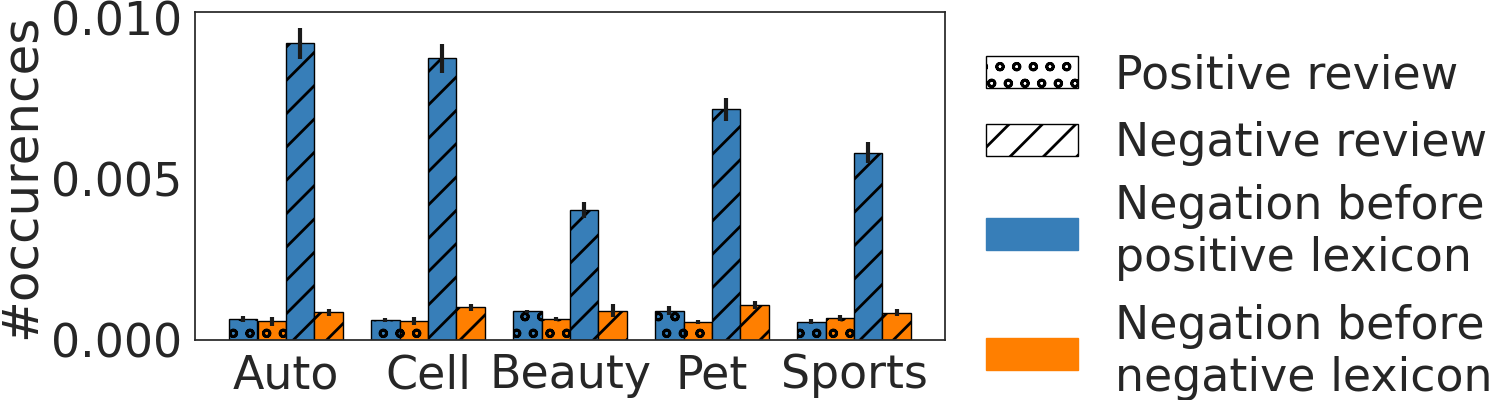}
  \caption{Amazon datasets.}
  \label{fig:vader_sentiment_negation_w_lvl_amz}
\end{subfigure}
\caption{Word-level negation distribution using Vader. \figref{fig:vader_overall_negation_w_lvl_non_amz} and \figref{fig:vader_overall_negation_w_lvl_amz} indicate the more frequent use of negation in negative reviews than in positive reviews at the word-level. Negative reviews have more negations before positive words in all datasets. This difference is substantially large in case of Yelp, Tripadvisor, PeerRead and Amazon reviews. This shows that although negative reviews have more positive words than negative words, these positive words are associated with negations.}
\label{fig:vader_negation_w_lvl_all}
\end{figure*}

\begin{figure*}
\centering
\begin{subfigure}[t]{.48\textwidth}
  \centering
  \includegraphics[width=0.95\textwidth]{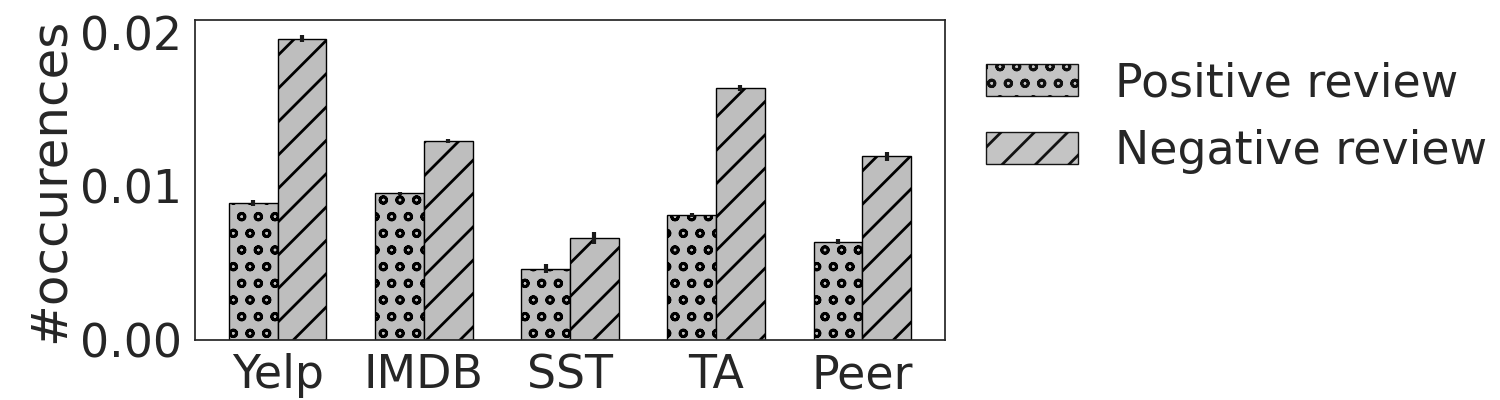}
  \caption{
  Overall negation in non-Amazon datasets.
  }
  \label{fig:dp_overall_negation_w_lvl_non_amz}
\end{subfigure}
\begin{subfigure}[t]{.48\textwidth}
  \centering
  \includegraphics[width=0.95\textwidth]{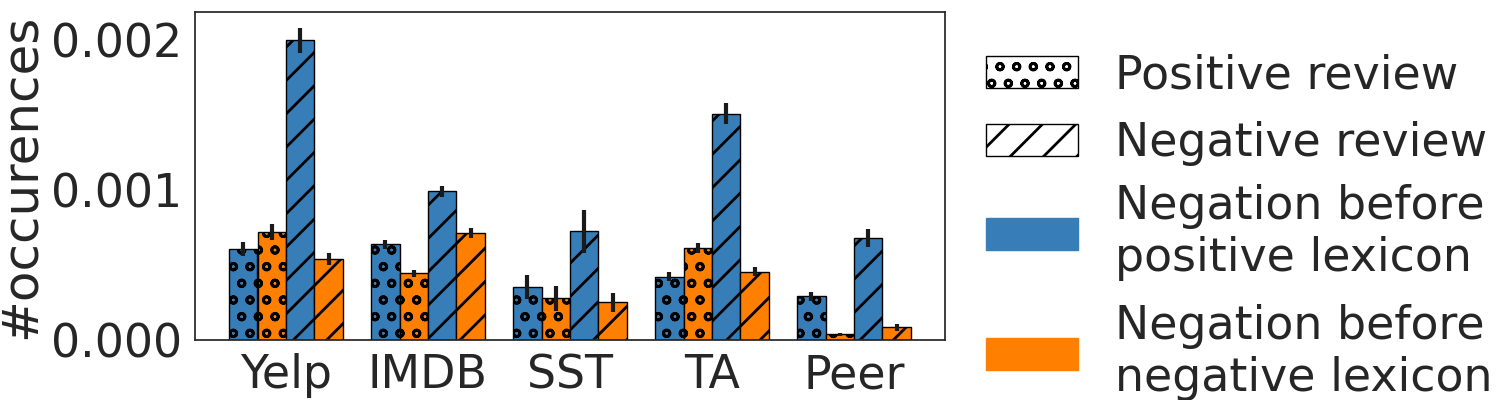}
  \caption{
  Negation before positive and negative lexicons in non-Amazon datasets.
  }
  \label{fig:dp_sentiment_negation_w_lvl_non_amz}
\end{subfigure}
\begin{subfigure}[t]{.48\textwidth}
  \centering
  \includegraphics[width=0.95\textwidth]{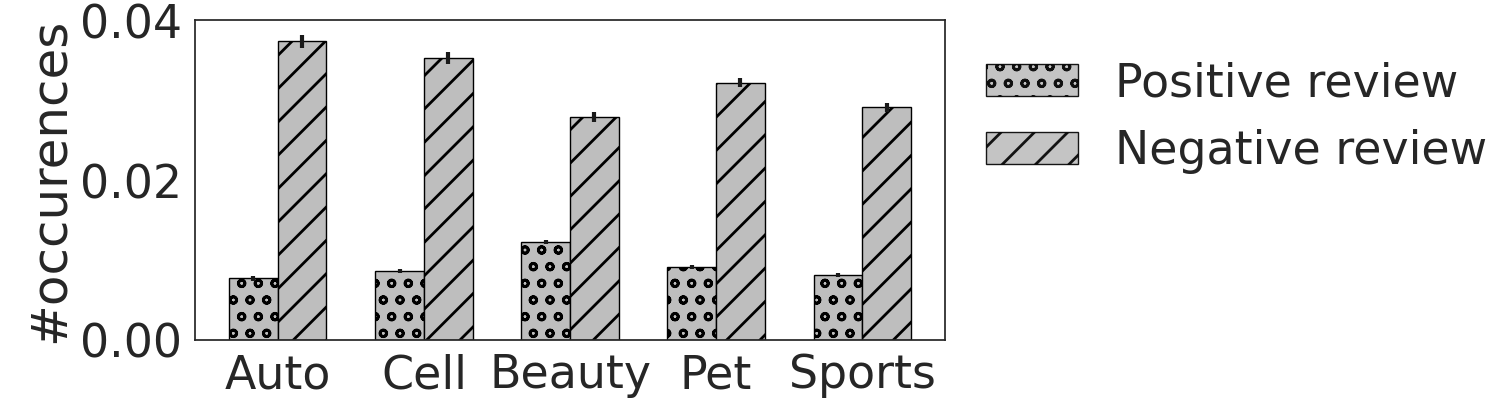}
  \caption{
  Overall negation in Amazon datasets.
  }
  \label{fig:dp_overall_negation_w_lvl_amz}
\end{subfigure}
\begin{subfigure}[t]{.48\textwidth}
  \centering
  \includegraphics[width=0.95\textwidth]{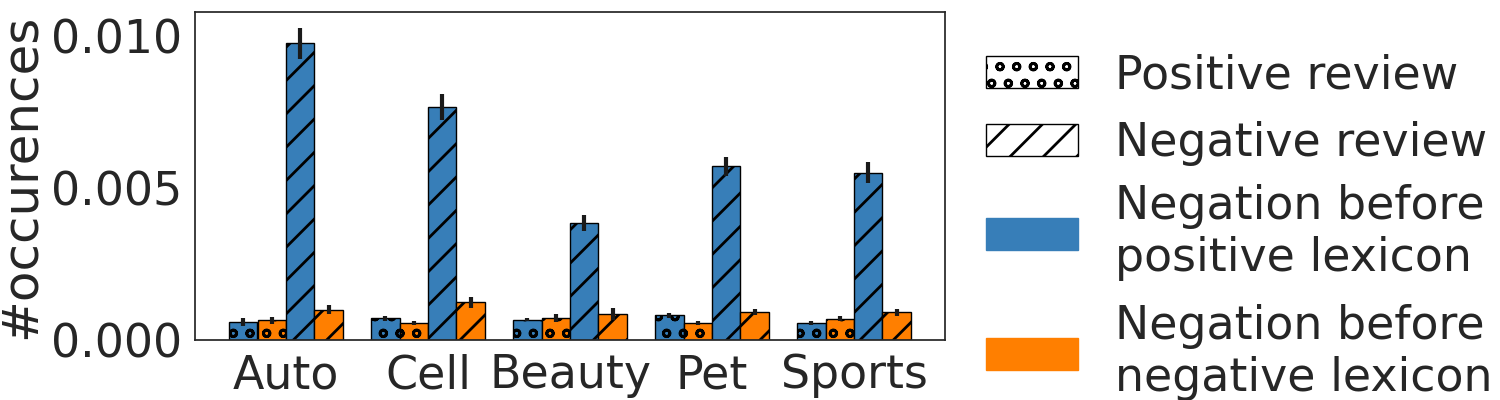}
  \caption{
  Negation before positive and negative lexicons in Amazon datasets.
  }
  \label{fig:dp_sentiment_negation_w_lvl_amz}
\end{subfigure}
\caption{Word-level negation distribution of all reviews using dependency parsing. 
With dependency parsing, we observe the same pattern as in \figref{fig:vader_negation_w_lvl_all}.
Negative reviews in Yelp, Tripadvisor, PeerRead and Amazon datasets have substantially more negations in general and also before positive words. This high number of negation associated with positive words can partially explain the higher proportion of positive words in negative reviews.
}
\label{fig:dp_negation_w_lvl_all}
\end{figure*}

\para{Word-level results}. \figref{fig:lexicon_dist_w_lvl_all} shows the lexicon distribution using LIWC and Vader. See \figref{fig:vader_negation_w_lvl_all} and \figref{fig:dp_negation_w_lvl_all} for word-level results of negation distribution using Vader and dependency parsing respectively.

\end{document}